\documentclass{article}
\usepackage[T1]{fontenc}
\usepackage{iclr2027_conference,times}
\usepackage{amsmath,amssymb,amsthm}
\usepackage{graphicx,booktabs,array,tabularx}
\usepackage{float}
\usepackage{xcolor}
\usepackage{url}
\usepackage[hidelinks]{hyperref}
\newtheorem{theorem}{Theorem}
\newtheorem{proposition}{Proposition}
\newtheorem{lemma}{Lemma}

\theoremstyle{definition}

\theoremstyle{plain}

\newcounter{algorithm}
\makeatletter
\newenvironment{algorithmblock}[1]{\par\medskip\refstepcounter{algorithm}
 \noindent\begin{minipage}{\linewidth}\hrule\vspace{5pt}
 \textbf{Algorithm \thealgorithm. #1}\par\vspace{2pt}\small
 \let\originalenumerate\enumerate
 \def\enumerate{\originalenumerate\setlength{\itemsep}{2pt}\setlength{\parsep}{0pt}\setlength{\topsep}{3pt}}
 }{\vspace{4pt}\hrule\end{minipage}\par\medskip}
\newcommand{\externalref}[3]{\@namedef{r@#1}{{#2}{}{#2}{#3}{}}}
\makeatother
\iclrfinalcopy
\title{Learning Unknown Oscillators\\from Bandit Feedback}
\author{Priyank Jayraj \qquad Poonam Goyal \qquad Navneet Goyal\\
\normalfont Birla Institute of Technology and Science, Pilani, India}
\hypersetup{pdftitle={Learning Unknown Oscillators from Bandit Feedback},pdfauthor={Priyank Jayraj, Poonam Goyal, Navneet Goyal},pdfsubject={Oscillator learning from selected feedback: regret guarantees and empirical comparisons}}
\begin{document}
\maketitle
\lhead{Preprint}
\newcommand{\RevCertZero}{3.36}
\newcommand{\RevSparseZero}{3.36}
\newcommand{\RevDenseZero}{2.99}
\newcommand{\RevTSZero}{4.26}
\newcommand{\RevCertMixed}{2.36}
\newcommand{\RevSparseMixed}{2.36}
\newcommand{\RevDenseMixed}{2.01}
\newcommand{\RevTSMixed}{2.36}
\newcommand{\RevPrimaryComparison}{With zero mixed observations, checked commitment reduces mean regret by 21.2\% relative to the posterior. The paired reduction is 0.90, with interval [0.46,1.36] adjusted for the two sensing comparisons. Startup contributes 1.18 of this gain; afterward the posterior recovers 0.27. Against dense NLS, holding predictions costs 0.37 regret (12.4\%) while using 38 fits instead of 1023. One mixed observation gives the posterior and checked controller essentially equal catalog means.}
\newcommand{\RevGateSummary}{All 1920 primary block checks pass, and the controller takes the same actions as its sparse NLS control in all 384 trajectories.}
\newcommand{\RevCertificateResults}{For the 48 zero-mixed trajectories, the mean upper bound is 3.97 and observed mean regret is 3.08. For the 48 mixed-start trajectories, these values are 3.32 and 2.48. Every reported bound exceeds its trajectory's measured regret.}

\newcommand{\RevConfirmationResults}{The shorter schedule has mean regret 3.30, compared with 3.25 for dense NLS, 3.55 for doubling, and 3.88 for the posterior. Its excess over dense NLS is 0.049 (95\% paired interval [0.024,0.072]), or 1.51\%, using 49 fits instead of 1023. Its mean regret is 15.0\% lower than the posterior (paired reduction 0.58; adjusted interval [0.06,1.11]). Both checked schedules pass every block test; the shorter schedule and its matched sparse control take identical actions in all 96 instances.}
\newcommand{\RevFreshCertResults}{The 24 shorter-block confirmation runs have mean bound 3.96 and observed mean regret 3.27.}
\newcommand{\RevRefitRows}{Dense NLS & 1023 & 1023 & Every nonempty history \\
Checked commitment & 38 & 49 & Proposed-block checkpoints \\
Matched sparse NLS & 38 & 49 & Prescribed checkpoints \\}
\newcommand{\RevRefitComparison}{Checked commitment uses 96.3\% fewer nonlinear refits than dense NLS on the primary catalog and 95.2\% fewer on the fresh catalog. Matched sparse NLS has the same counts, so this comparison isolates fitting cadence rather than certification. Appendix~\ref{app:runtime} reports isolated runtimes including confidence computation.}
\newcommand{\RevTwoCert}{11.95}
\newcommand{\RevTwoNLS}{11.95}
\newcommand{\RevTwoPlaneResults}{All 80 primary checks pass, and the two policies take identical actions in all sixteen instances. Certification adds prospective reward and information bounds to those decisions. Terminal trajectory bounds range from 47.06 to 65.40, with the failure budget divided across the sixteen instances for simultaneous coverage.}
\newcommand{\RevStressResults}{Starting the two-plane checks at round 4 causes four refusals and sixty online-learning rounds in each run. Both first hold a model after round 64. Final regrets 74.23 and 67.29 fall below trajectory bounds 119.01 and 109.37.}

\begin{abstract}
We study semi-bandits driven by unknown oscillators. Each round, a learner selects two of four reward channels and observes their individual rewards. Frequency errors change future reward rankings and the resulting measurements. We show when committing to one fitted oscillator lets its greedy choices expose prediction errors. Our controller bounds reward loss (regret) over a continuous confidence set and holds the longest block within budget. An online regression policy gathers observations when the check fails. Regret grows at a square-root rate, up to logarithmic factors, across all frequencies and extends to two competing oscillators. On 96 fresh synthetic instances, checked commitment uses 49 fits instead of 1,023, with 1.5\% higher mean regret than fitting every round. Matched isolated timings give a 4.69-fold speedup. The confidence sets also certify trajectory regret.
\end{abstract}

\section{Introduction}
\label{sec:intro}
A decision maker chooses two reward channels from four whose values rise and fall over time. Two channels follow the horizontal coordinate of a rotating signal; two follow its vertical coordinate. Only the chosen rewards are observed. Choosing both copies of the better coordinate earns the largest immediate reward, but reads the same part of the signal twice. The decision that earns reward also determines what the learner can discover.

\begin{figure}[!h]
\centering
\includegraphics[width=\linewidth]{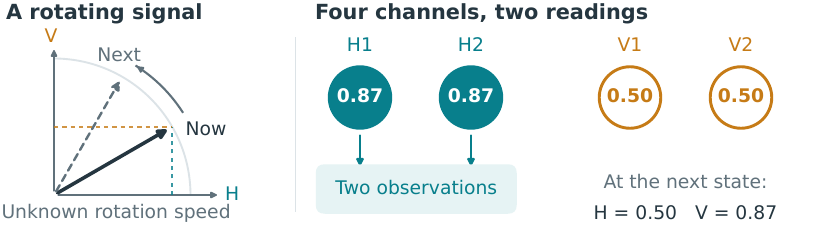}
\caption{Two measurements of a rotating signal. H and V denote horizontal and vertical coordinates; subscripts distinguish their two equal-mean channels. Selecting H1 and H2 reads one coordinate twice. Rotation changes which pair earns more.}
\label{fig:task}
\end{figure}

Recurring network demand motivates similar allocation questions \citep{oh2019periodic}. Here the unknown signal is an \emph{oscillator}: a state moving around a circle at a fixed speed. Frequency is its rotation speed, and phase is its current position. Even a small speed error eventually puts a forecast in the wrong part of the cycle. Learning therefore requires both a useful prediction and measurements that can expose its mistakes. Figure~\ref{fig:task} gives the complete sensing geometry.

Our starting observation is that a model can choose its own test. A \emph{greedy} decision takes the pair with the largest predicted reward. Under \emph{model commitment}, one fitted oscillator supplies these decisions for a block of rounds. The chosen coordinate changes as its prediction rotates. Keeping the same explanation can therefore produce a revealing measurement that repeated changes of explanation avoid.

\clearpage

The central question is how long one model should guide the observations.
Our geometric analysis first relates agreement on the chosen coordinates
to agreement on the underlying signal. It shows how holding a model can
make its prediction errors visible, and quantifies the connection between
errors on selected coordinates and future forecasts. 

We turn duration into a decision that can be checked from past rewards.
A fitted oscillator proposes future choices. A confidence set retains
the models supported by the observations, and the controller bounds the
reward those choices could lose under any retained model. It holds the
longest block within a specified budget. When more information is needed,
an online learner supplies the next observations. The resulting guarantee
covers every rotation speed and extends to two oscillators competing for
the same two measurements.

We measure performance through \emph{regret}, the accumulated reward lost
against the best pair in each round. The analysis gives a square-root
regret guarantee and bounds computed for complete observed trajectories.
Matched experiments then measure how the duration of commitment affects
reward, refitting work, and elapsed time. 

\section{Related work}
\label{sec:related}

\paragraph{Information from greedy choices.}
Reward-seeking decisions can generate the information needed for learning.
In contextual bandits, feature diversity and smoothed analyses give
conditions under which greedy choices remain informative
\citep{bastani2021greedy,kannan2018smoothed,raghavan2018diversity}.
Greedy-First monitors whether the collected information suffices
\citep{bastani2021greedy}.
For finite structured reward classes, self-identifiability characterizes
asymptotic greedy success through observations that reveal a policy's suboptimality
\citep{slivkins2025greedy}. Optimistic and pessimistic hypotheses also
support directed sampling \citep{jun2020crop}.
The oscillator geometry lets us study how errors become visible along
the coordinate sequence chosen by that same model. The resulting
geometric bounds connect selected discrepancies to prediction
error across time, including the ambiguity between different frequencies.

\paragraph{Recurring rewards and hidden dynamics.}
An unobserved oscillator continues to evolve, as do the arms in restless
bandit models \citep{whittle1988restless}.
For unknown irreducible finite-state Markov arms, \citet{ortner2012restless}
give square-root regret up to logarithmic factors, relative to a policy
that knows the transition laws and reward means.
Algorithms also compete with periodic action patterns
\citep{oh2019periodic}, while seasonal bandits estimate unknown integer
periods and combine observations from corresponding phases
\citep{chen2025seasonal}. A continuous rotation describes a shared signal
whose sampled values need not repeat after an integer number of rounds.
Sliding-window methods estimate recent reward levels
\citep{garivier2011ucb}. They offer a complementary way to use changing
rewards; the oscillator model connects separated observations through
the same underlying dynamics.
Classical frequency estimation recovers sinusoidal parameters from
discrete or nonuniform samples
\citep{rife1974single,abbas2016nonuniform,davis2019frequency}.
Here the estimated signal also determines which coordinates are measured.
For known linear Gaussian systems, \citet{gornet2025kode} connect greedy
decisions with observability, the ability to recover a hidden state
from its measurements. \citet{gornet2024restless} learn reward predictors
from selected histories under unknown dynamics. Our setting exposes
the role of an unknown rotation speed and the frequency ambiguities
created by selected coordinate measurements.

\paragraph{Regret and confidence from selected data.}
Selecting several arms and observing their individual rewards gives
combinatorial semi-bandit feedback \citep{kveton2015tight}.
SquareCB converts an online regression guarantee into a regret guarantee
\citep{foster2020beyond}; it supplies our learning decisions when a
proposed block requires more information. Sequential likelihood methods
provide confidence sets valid under adaptive data collection
\citep{wasserman2020universal,emmenegger2023likelihood}.
We evaluate future decision costs over a continuous oscillator confidence
set, using the quadratic state dependence at each frequency and interval
bounds across frequencies. This also yields bounds for an observed
trajectory. Thompson sampling draws a model from a distribution updated with observed rewards and chooses its preferred action \citep{russo2018tutorial}. Our oscillator
posterior comparator shares information across coordinates and updates
every round. 

\clearpage
\section{A choice is also a measurement}
\label{sec:model}\label{sec:learn}
Rounds are indexed by $t=1,\ldots,T$, where $T$ is the horizon. The unknown initial state is $z\in\mathbb R^2$, and the unknown frequency $\omega$ is the rotation angle per round. Let $R(\psi)$ denote counterclockwise rotation by angle $\psi$. The state in round $t$ is $R(t\omega)z$.

There are four physical arms. A known label $b(i)\in\{1,2\}$ assigns each arm to a coordinate, with two arms per coordinate. These equal-mean arms are \emph{replicas}. With coordinate basis vectors $e_1,e_2$ and observation noise $\varepsilon_{i,t}$, the means and observations are
\[
 \mu_i(t)=e_{b(i)}^\top R(t\omega)z,
 \qquad Y_{i,t}=\mu_i(t)+\varepsilon_{i,t}.
\]
Before observing current rewards, the learner chooses two distinct arms $S_t$. It sees only $Y_{i,t}$ for $i\in S_t$. Conditional on the past, selected noise is centered and jointly $\sigma$-sub-Gaussian: every linear combination obeys the conditional moment bound in Appendix~\ref{sec:gaussian-mixture-confidence}. The experiments use independent Gaussian noise.

Let $\mathcal F$ be the six possible pairs. We measure the accumulated mean reward lost against the best pair each round, called \emph{pseudo-regret}:
\[
 R_T=\sum_{t=1}^T\left\{
 \max_{A\in\mathcal F}\sum_{i\in A}\mu_i(t)-\sum_{i\in S_t}\mu_i(t)\right\}.
\]
The amplitude $\|z\|_2$ measures signal strength. With a supplied bound $A_{\max}>0$, the controller learns over the continuous class
\[
 \Theta=\{(\omega,z):\omega\in[0,2\pi),\ \|z\|_2\le A_{\max}\}.
\]
The learner receives the labels, $A_{\max}$, $\sigma$, $T$, and a failure probability $\delta\in(0,1)$. Frequency and initial state remain fixed. This class includes zero amplitude and every rotation speed.

An optimal pair takes both replicas of the larger coordinate. We call the horizontal and vertical pairs H and V. A \emph{mixed pair} takes one arm from each coordinate. Replication makes the allocation tension explicit: using both slots on the attractive coordinate gives two noisy readings, while spreading them reveals both coordinates. Section~\ref{sec:twoplane} keeps the same observation budget and replaces the replicas with four distinct coordinates of two signals.

\paragraph{An explanation that avoids its own test.}
A mixed observation reveals a noisy state vector, but learning its rotation speed still requires measurements at different times. We use the following noiseless example to isolate that distinction. An \emph{alias} is a different model compatible with the observations. Set $z=e_1$. Rotations $2\pi/3$ and $-2\pi/3$ have identical horizontal coordinates and opposite vertical coordinates. Take the positive rotation as truth. A switcher can use the reflected model in round 1 and the true model in round 2, choosing H both times. Both models choose H in round 3. Repeating this pattern preserves the ambiguity.

\begin{table}[!ht]
\centering\small
\caption{Holding one explanation changes the selected measurement.}
\label{tab:commit}
\begin{tabular*}{\linewidth}{@{\extracolsep{\fill}}llll@{}}
\toprule
\textbf{Rule} & \textbf{Round 1} & \textbf{Round 2} & \textbf{Consequence}\\
\midrule
Specified switcher & Reflection; H & Truth; H & Alias survives\\
Hold reflection & Reflection; H & Reflection; V & Alias rejected\\
\bottomrule
\end{tabular*}
\end{table}
\begin{proposition}[A held model exposes its alias]
\label{prop:alias-main}
In the noiseless construction of Table~\ref{tab:commit}, the specified switcher incurs $(1+\sqrt3)\lceil T/3\rceil$ regret. Holding the reflection until the first inconsistent observation and then selecting the truth incurs at most $2\sqrt3$ regret.
\end{proposition}

The held reflection predicts $+\sqrt3/2$ on V in round 2, where the observation is $-\sqrt3/2$. Its greedy choice exposes the error. The next result extends this mechanism from exact aliases to quantitative prediction control.

\clearpage
\section{What a held model reveals}
\label{sec:geometry}
The relevant information is the disagreement on the measurements a model chooses. On a block of $n$ consecutive rounds, measure time from its start. Write $(\omega,z)$ for the truth and $(\nu,y)$ for a held candidate. If $S_t$ is greedy for that same candidate on every round, define its selected squared discrepancy by
\[
 D_n=\sum_{t=1}^n\sum_{i\in S_t}
 \left[e_{b(i)}^\top\{R(t\omega)z-R(t\nu)y\}\right]^2.
\]
This compares means, before observation noise. Let $R_{1:n}$ denote regret on the held block. Define
$\lambda(\alpha)=1-\max\{|\cos\alpha|,|\sin\alpha|\}$.
A positive floor on $\lambda$ keeps the rotation away from multiples of a quarter turn, where successive measurements can repeatedly miss the same direction. We call these speeds \emph{resonances}.

\begin{theorem}[Prediction control from committed measurements]
\label{thm:committed-prediction}
For every $0<\lambda_0\le1-1/\sqrt2$, there are finite constants $H\ge2$ and $K$, depending only on $\lambda_0$, such that the following holds. Let $z,y\in\mathbb R^2$ be arbitrary and let $\lambda(\omega),\lambda(\nu)\ge\lambda_0$. For every integer $n\ge H$ and every greedy tie rule of the held candidate,
\[
 \max_{t\in\{1,\ldots,2n\}}
 \|R(t\omega)z-R(t\nu)y\|_2^2\le K\frac{D_n}{n},
 \qquad R_{1:n}\le4\sqrt{K nD_n}.
\]
Both states may have zero amplitude.
\end{theorem}

Agreement on the selected coordinates controls the whole state for a further block of the same length. Arbitrary coordinate histories can preserve reflected aliases; requiring one model to drive the choices removes that freedom. The proof first establishes finite identifiability under greedy choices, then turns it into a uniform error bound. Consecutive sensing directions control state error; persistence across longer blocks distinguishes rotation speeds. A scale argument extends the bound to arbitrarily small amplitudes (Appendix~\ref{app:committed-prediction}). The conclusion concerns predictions: a vanishing signal can have an uncertain frequency while its reward forecast remains accurate.

\section{Fit, check, and hold}
\label{sec:oneplane}\label{sec:info}
The structural result explains why persistence can supply information. The controller decides its duration directly from observations, over the full class $\Theta$.

\paragraph{Retain the models supported by the data.}
Model $\theta=(\nu,x)$ predicts $\mu_i^\theta(t)=e_{b(i)}^\top R(t\nu)x$. After $n$ rounds, its selected squared loss is
\[
 L_n(\theta)=\sum_{t\le n}\sum_{i\in S_t}(Y_{i,t}-\mu_i^\theta(t))^2.
\]
Nonlinear least squares (NLS) proposes a fit $\widehat\theta_n$. A sequential statistical test defines a \emph{confidence set} $\mathcal C_n$ that contains the truth simultaneously at decision times with the declared probability.

\paragraph{Price the proposed decisions.}
After a startup of $1\le n_0\le T$ rounds, use growth factor $q>1$ to propose the fit's greedy pairs $S_t^{(n)}$ through $e=\min\{T,\lceil qn\rceil\}$. Their largest compatible regret is
\[
 U_{n,e}=\sup_{\theta\in\mathcal C_n}\sum_{t=n+1}^{e}
 \left\{\max_{A\in\mathcal F}\sum_{i\in A}\mu_i^\theta(t)
       -\sum_{i\in S_t^{(n)}}\mu_i^\theta(t)\right\}.
\]
With a computed upper bound $\overline U_{n,e}$ and budget coefficient $c>0$, accept a prefix when
\begin{equation}\label{eq:gate-main}
 \overline U_{n,e}\le c\sigma\sqrt{\log(8T/\delta)}(\sqrt e-\sqrt n).
\end{equation}

\clearpage
\section{A complete learning rule and its guarantee}
\label{sec:guarantees}
The same confidence calculation bounds every integer prefix of a proposed block. This lets the controller keep useful parts of a forecast when the full duration is too costly. If no prefix passes, an online regression policy gathers further information.

\begin{algorithmblock}{Checked model commitment for one oscillator}
\label{alg:reconstructed}
\textbf{Input:} $\Theta,T,\sigma,\delta,n_0,q,c$, an online learner, and fixed tie orders. Split $\delta$ equally between confidence coverage and online learning.
\begin{enumerate}
\item Execute the declared startup through $n_0$, storing selected rewards.
\item At $n$, fit $\widehat\theta_n$ and bound its greedy schedule through $e=\min\{T,\lceil qn\rceil\}$. Test prefixes using cumulative per-round bounds.
\item Execute the longest prefix passing \eqref{eq:gate-main}. If none passes, run SquareCB for $\min\{4,T-n\}$ rounds. This online policy randomizes toward promising pairs and updates its regressors from the resulting selected rewards.
\item Store observations, set $n$ to the actual endpoint, and repeat through $T$. An unresolved check selects the learning route.
\end{enumerate}
\end{algorithmblock}

\begin{theorem}[A full-circle regret guarantee]
\label{thm:reconstructed-main}
For the one- or two-oscillator model, let $G=2\sqrt2 A_{\max}$ bound one round's regret. Let $F_T=\widetilde O(\sqrt T)$ be the online-learning bound in \eqref{eq:recovery-finite-coefficient}; the tilde suppresses logarithmic factors in $T$ and $1/\delta$. Algorithm~\ref{alg:reconstructed} and its two-plane extension, with the online learner of Appendix~\ref{app:active-rate}, satisfy with probability at least $1-\delta$,
\[
 R_T\le G n_0+c\sigma\sqrt{\log(8T/\delta)}
 (\sqrt T-\sqrt{n_0})+F_T.
\]
Any increasing checkpoints and any startup rule based on information available before each round are allowed. The actual startup regret may replace $Gn_0$, and $GT$ also bounds the total.
\end{theorem}

Accepted budgets telescope because each block receives the increment $\sqrt e-\sqrt n$. On learning rounds, SquareCB converts prediction error into regret. For each trial frequency, a two-dimensional linear learner estimates the initial state. Aggregation over a sufficiently fine full-circle cover controls prediction error, including at resonances. Combining the accepted blocks and learning rounds proves the theorem; Appendix~\ref{app:active-rate} gives the cover, explicit coefficients, and probability allocation.

\begin{proposition}[A phase uncertainty floor]
\label{prop:phasefloor}
For every amplitude $A>0$, noise scale $\sigma\ge0$, and horizon $T\ge1$, two models $\theta_+,\theta_-$ with different phases at frequency $\pi/4$ satisfy, under independent Gaussian noise of variance $\sigma^2$,
\[
 \max_{\theta\in\{\theta_+,\theta_-\}}\mathbb E_\theta R_T
 \ge\tfrac1{64}\min\{AT,\sigma\sqrt T\}.
\]
This holds for every policy, including one given the frequency and amplitude and allowed mixed pairs in every round.
\end{proposition}

\section{Computing useful confidence bounds}
\label{sec:implementation}\label{sec:certified}
At a fixed frequency, selected loss is quadratic in the initial state. Define the sensing column $d_{i,t}(\nu)=R(-t\nu)e_{b(i)}$ and, summing over observed rewards, let
\[
 C_n=\sum Y_{i,t}^2,\quad b_n(\nu)=\sum Y_{i,t}d_{i,t}(\nu),\quad
 Q_n(\nu)=\sum d_{i,t}(\nu)d_{i,t}(\nu)^\top.
\]
Then $L_n(\nu,x)=C_n-2b_n(\nu)^\top x+x^\top Q_n(\nu)x$.
For $\sigma>0$, a uniform frequency prior and an isotropic Gaussian state prior supply a sequential exponential loss test. We intersect its confidence set with $\Theta$. Integrating the two-dimensional state analytically reduces confidence construction to a frequency calculation. Whole-interval bounds retain every frequency interval that can contain a compatible state. Numerical fits guide refinement; continuous bounds determine the confidence set used by the decision rule.

\clearpage
\section{From decision bounds to two competing signals}
\label{sec:twoplane}
The confidence calculation serves two purposes. Before a block, it bounds the cost of the planned choices. After a run, it bounds the loss of the complete action sequence using only the selected data.

\paragraph{Certify an observed trajectory.}
For actual actions $S_1,\ldots,S_T$, let $\mathcal R_T(\theta)$ be their pseudo-regret evaluated under model $\theta$. Construct $\mathcal C_T$ with failure budget $\delta_c$ and compute
\[
 \mathcal B_T\ge\sup_{\theta\in\mathcal C_T}\mathcal R_T(\theta).
 \qquad\text{Then}\qquad
 \Pr\{R_T\le\mathcal B_T\}\ge1-\delta_c.
\]
Each retained frequency interval encloses its compatible states by a ball and then a circumscribed polygon. At fixed frequency, regret is convex in the state, so its maximum occurs at a polygon vertex. A frequency remainder covers the whole interval. Prospective blocks instead use regularized ellipsoids to bound each possible action advantage. Adding these bounds prices every prefix in \eqref{eq:gate-main}. Appendix~\ref{app:structural-current-numerics} specifies the interval refinement and arithmetic bounds.

\paragraph{Allocate measurements between two oscillators.}
Plane $j\in\{1,2\}$ has state $x_j$, frequency $\omega_j$, and means
\[
 \mu_{j,b}(t)=e_b^\top R(t\omega_j)x_j,
 \qquad b\in\{1,2\},\quad \|x_j\|_2\le A_{\max}.
\]
There is one arm per plane and coordinate: four distinct arms and six possible pairs. Both frequencies range over the full circle. The labels are known. A good pair can use both measurements on one plane or split them across planes. The new difficulty is that a whole signal can receive no observation for several rounds.

\paragraph{Greedy allocation can reveal both planes.}
For positive-amplitude planes, accurate predictions and sufficient phase spread make greedy top-two selection informative in every state direction (Lemma~\ref{struct:lem:spatial}). The result uses the phase coverage of each plane separately, so equal frequencies are allowed. Its proof identifies rounds when a plane must enter the chosen pair and the observed coordinate has a substantial projection on a given direction. Appendix~\ref{struct:app:twoplane} gives the phase-spread condition and proof.

For the executed controller, the fitted oscillators propose a greedy schedule through $e=\min\{2n,T\}$. Write $\theta=((\omega_1,x_1),(\omega_2,x_2))$ for a joint model. For plane confidence sets $\mathcal C_{1,n}$ and $\mathcal C_{2,n}$, use the joint set $\mathcal C_n=\mathcal C_{1,n}\times\mathcal C_{2,n}$ to bound block regret. Two further checks evaluate what the proposed measurements earn and reveal.

\paragraph{Compare the cost of dedicated sensing.}
Let $F_t$ be a specified sensing schedule on the same next $m$ rounds as a proposed schedule $S_t$. Define their reward difference under a candidate model by
\[
 G_n(\theta)=\sum_{t=n+1}^{n+m}
 \left\{\sum_{i\in S_t}\mu_i^\theta(t)-\sum_{i\in F_t}\mu_i^\theta(t)\right\}.
\]
A positive lower bound on $\inf_{\theta\in\mathcal C_n}G_n(\theta)$ certifies the reward saved by using the proposal instead of that sensing schedule.

\paragraph{Check the planned information.}
For each plane, the planned sensing columns are $d_{j,t,b}(\nu)=R(-t\nu)e_b$. Sum their outer products over selected arms to form $Q_j^{\rm plan}(\nu)$. Its smallest eigenvalue measures information in the least-observed state direction. A lower bound over all compatible frequencies checks that both planes receive directional information. This calculation applies to the actual schedule, including equal or near-resonant frequencies.

The controller holds the full block when all checks pass. Otherwise it runs the online learner over all six pairs through $e$. Separate online regressors for each plane and observation slot avoid a product search over frequency grids (Appendix~\ref{app:active-rate}). They use the original round indices, so an unobserved plane keeps evolving. Additional checks only reduce the accepted blocks; the regret guarantee of Theorem~\ref{thm:reconstructed-main} follows from the same budget and learning terms.

\newcommand{\RevStressFreshPreviousCount}{32}
\newcommand{\RevStressFreshPreviousCheckedRegret}{4.070}
\newcommand{\RevStressFreshPreviousDenseRegret}{3.403}
\newcommand{\RevStressFreshPreviousPairedDifference}{0.666}
\newcommand{\RevStressFreshPreviousCheckedFits}{75.31}
\newcommand{\RevStressFreshPreviousDenseFits}{1023.00}
\newcommand{\RevStressFreshPreviousTotalCheckedFits}{2410}
\newcommand{\RevStressFreshPreviousTotalDenseFits}{32736}
\newcommand{\RevStressFreshPreviousFallbackRounds}{4238}
\newcommand{\RevStressFreshPreviousShortenedBlocks}{223}
\newcommand{\RevStressFreshHoldoutCount}{16}
\newcommand{\RevStressFreshHoldoutCheckedRegret}{3.845}
\newcommand{\RevStressFreshHoldoutDenseRegret}{2.590}
\newcommand{\RevStressFreshHoldoutPairedDifference}{1.255}
\newcommand{\RevStressFreshHoldoutCheckedFits}{77.44}
\newcommand{\RevStressFreshHoldoutDenseFits}{1023.00}
\newcommand{\RevStressFreshHoldoutTotalCheckedFits}{1239}
\newcommand{\RevStressFreshHoldoutTotalDenseFits}{16368}
\newcommand{\RevStressFreshHoldoutFallbackRounds}{2148}
\newcommand{\RevStressFreshHoldoutShortenedBlocks}{136}
\newcommand{\RevStressFreshRows}{$0$ & 0.15 & 1.950 & 0.675 & 1.350 & 0.750 & 8 & 50 \\
$0$ & 1.20 & 0.000 & 0.000 & 0.000 & 0.000 & 1980 & 0 \\
$0.01$ & 0.15 & 9.607 & 8.548 & 10.844 & 7.157 & 156 & 31 \\
$0.01$ & 1.20 & 1.641 & 0.943 & 1.758 & 0.435 & 4 & 2 \\
$\pi/2$ & 0.15 & 0.600 & 0.600 & 0.600 & 0.600 & 0 & 2 \\
$\pi/2$ & 1.20 & 5.940 & 5.940 & 3.394 & 3.394 & 0 & 32 \\
$\pi/2+0.01$ & 0.15 & 7.240 & 5.057 & 7.031 & 4.395 & 0 & 19 \\
$\pi/2+0.01$ & 1.20 & 5.577 & 5.465 & 5.785 & 3.991 & 0 & 0 \\}
\newcommand{\RevStressFreshAggregate}{Across the 32 previously inspected instances, mean cumulative regret is 4.070 for the checked controller and 3.403 for dense NLS (paired checked-minus-dense mean 0.666). Across the 16 held-out instances, the corresponding means are 3.845 and 2.590 (paired difference 1.255). The checked controller uses 75.31 and 77.44 nonlinear refits per instance in the previous and held-out cohorts, respectively; dense NLS uses 1023.00 and 1023.00. The previous cohort has 4238 fallback rounds and 223 shortened blocks in total; the held-out cohort has 2148 and 136, respectively.}

\newcommand{\ReviewStressPreviousCheckedMean}{4.070}
\newcommand{\ReviewStressPreviousCheckedMaximum}{13.22}
\newcommand{\ReviewStressPreviousCheckedFits}{75.31}
\newcommand{\ReviewStressPreviousSparseMean}{5.117}
\newcommand{\ReviewStressPreviousSparseMaximum}{16.65}
\newcommand{\ReviewStressPreviousSparseFits}{38}
\newcommand{\ReviewStressPreviousDenseMean}{3.403}
\newcommand{\ReviewStressPreviousDenseMaximum}{15.72}
\newcommand{\ReviewStressPreviousDenseFits}{1023}
\newcommand{\ReviewStressPreviousPairedMean}{-1.048}
\newcommand{\ReviewStressPreviousWinTieLoss}{9/16/7}
\newcommand{\ReviewStressPreviousLearningRounds}{4238}
\newcommand{\ReviewStressPreviousShortenedBlocks}{223}
\newcommand{\ReviewStressHoldoutCheckedMean}{3.845}
\newcommand{\ReviewStressHoldoutCheckedMaximum}{14.61}
\newcommand{\ReviewStressHoldoutCheckedFits}{77.44}
\newcommand{\ReviewStressHoldoutSparseMean}{6.209}
\newcommand{\ReviewStressHoldoutSparseMaximum}{33.55}
\newcommand{\ReviewStressHoldoutSparseFits}{38}
\newcommand{\ReviewStressHoldoutDenseMean}{2.590}
\newcommand{\ReviewStressHoldoutDenseMaximum}{9.01}
\newcommand{\ReviewStressHoldoutDenseFits}{1023}
\newcommand{\ReviewStressHoldoutPairedMean}{-2.364}
\newcommand{\ReviewStressHoldoutWinTieLoss}{4/8/4}
\newcommand{\ReviewStressHoldoutLearningRounds}{2148}
\newcommand{\ReviewStressHoldoutShortenedBlocks}{136}
\newcommand{\ReviewStressAllLearningRounds}{6386}
\newcommand{\ReviewStressAllShortenedBlocks}{359}
\newcommand{\ReviewStressRows}{
R01 & $0$ & 0.15 & 1.950 & 2.100 & 0.675 & 1.350 & 6.450 & 0.750 & 8 & 50 \\
R02 & $0$ & 1.20 & 0.000 & 0.000 & 0.000 & 0.000 & 0.000 & 0.000 & 1980 & 0 \\
R03 & $0.01$ & 0.15 & 9.607 & 14.283 & 8.548 & 10.844 & 22.427 & 7.157 & 156 & 31 \\
R04 & $0.01$ & 1.20 & 1.641 & 1.345 & 0.943 & 1.758 & 0.514 & 0.435 & 4 & 2 \\
R05 & $\pi/2$ & 0.15 & 0.600 & 0.600 & 0.600 & 0.600 & 0.600 & 0.600 & 0 & 2 \\
R06 & $\pi/2$ & 1.20 & 5.940 & 5.940 & 5.940 & 3.394 & 3.394 & 3.394 & 0 & 32 \\
R07 & $\pi/2+0.01$ & 0.15 & 7.240 & 11.094 & 5.057 & 7.031 & 10.502 & 4.395 & 0 & 19 \\
R08 & $\pi/2+0.01$ & 1.20 & 5.577 & 5.577 & 5.465 & 5.785 & 5.785 & 3.991 & 0 & 0 \\
\midrule
\multicolumn{3}{l}{Mean} & 4.070 & 5.117 & 3.403 & 3.845 & 6.209 & 2.590 & -- & -- \\
\multicolumn{3}{l}{Median} & 3.203 & 3.590 & 2.225 & 3.043 & 3.394 & 1.653 & -- & -- \\
\multicolumn{3}{l}{90th percentile} & 9.023 & 12.723 & 7.432 & 8.411 & 12.361 & 5.843 & -- & -- \\
\multicolumn{3}{l}{Maximum} & 13.224 & 16.652 & 15.716 & 14.605 & 33.545 & 9.013 & -- & -- \\
\multicolumn{3}{l}{Mean refits} & 75.31 & 38.00 & 1023.00 & 77.44 & 38.00 & 1023.00 & -- & -- \\
}
\newcommand{\ReviewResourceRows}{
R01 & 4 & 195.919 & 472.5 & 0.195 & 65.2 & 7.121 & 66.0 \\
R01 & 5 & 471.069 & 895.1 & 0.209 & 65.4 & 7.085 & 65.9 \\
R02 & 4 & 1295.072 & 5075.1 & 0.333 & 65.4 & 9.407 & 66.0 \\
R02 & 5 & 1327.704 & 5069.6 & 0.213 & 65.4 & 9.004 & 66.0 \\
R03 & 4 & 250.153 & 698.5 & 0.181 & 65.4 & 6.644 & 66.0 \\
R03 & 5 & 429.804 & 997.2 & 0.176 & 65.5 & 7.094 & 66.0 \\
R04 & 4 & 13.332 & 155.2 & 0.233 & 65.5 & 6.639 & 66.0 \\
R04 & 5 & 6.475 & 120.6 & 0.238 & 65.5 & 6.481 & 65.9 \\
R05 & 4 & 6.513 & 120.2 & 0.230 & 65.4 & 7.172 & 66.0 \\
R05 & 5 & 6.569 & 119.4 & 0.180 & 65.4 & 6.735 & 66.1 \\
R06 & 4 & 219.636 & 512.1 & 0.198 & 65.5 & 6.342 & 66.1 \\
R06 & 5 & 184.177 & 510.5 & 0.197 & 65.5 & 6.536 & 66.1 \\
R07 & 4 & 90.877 & 350.4 & 0.176 & 65.5 & 6.972 & 65.9 \\
R07 & 5 & 113.067 & 404.4 & 0.169 & 65.4 & 7.077 & 66.0 \\
R08 & 4 & 0.661 & 71.9 & 0.188 & 65.4 & 7.330 & 66.0 \\
R08 & 5 & 0.730 & 71.6 & 0.248 & 65.5 & 7.603 & 66.0 \\
\midrule
Two planes & 91001 & 78.529 & 575.6 & 0.407 & 75.7 & -- & -- \\
Two planes & 91002 & 75.233 & 565.4 & 0.393 & 76.0 & -- & -- \\
}
\newcommand{\ReviewResourceSummaryRows}{
Stress & $n=16$ & 288.235 & 5075.1 & 0.210 & 65.5 & 7.203 & 66.1 \\
Two planes & $n=2$ & 76.881 & 575.6 & 0.400 & 76.0 & -- & -- \\
}
\newcommand{\ReviewEarlyControlComparison}{Early two-plane checked and matched regrets are $74.23$ and $23.28$ at seed 91001, and $67.29$ and $10.96$ at seed 91002. All four runs use 11 joint refits (22 per-plane NLS solves).}

\clearpage
\section{Experiments}
\label{sec:experiments}\label{sec:exp-global}\label{sec:exp-selection}
We measure how long a fitted oscillator can guide decisions while retaining useful reward and confidence bounds. Tables~\ref{tab:structured-continuum} and~\ref{tab:confirmation-computation} vary the initial measurements, the time between fits, and the checks applied to a proposed schedule. All reported regret includes startup. Paired policies share Gaussian potential rewards indexed by physical arm and round; each learner sees only its selected entries.

\subsection{Learning from the same observation budget}
The primary catalog has twelve frequency, amplitude, and noise combinations spread around the frequency circle. Amplitudes are $1.021,1.247,1.479$, and noise levels are $0.05,0.10,0.20$. Sixteen seeds per cell give 192 paired instances at horizon $T=1024$. We compare two initial sensing rules. A \emph{pure start} selects replica pairs throughout. A \emph{mixed start} reads both coordinates in round one, then selects replica pairs.

The NLS proposals search the full frequency circle and amplitude disk. Checked commitment refits during the first 34 rounds and then proposes doubling blocks. Its matched sparse control uses the same fitting routine and nominal checkpoints. Dense NLS refits after every nonempty history. The independent Thompson sampler updates a posterior over the oscillator and draws a new model each round.

\begin{table}[!ht]
\centering\small
\caption{Primary mean regret over 192 paired instances per sensing rule. Bold marks the lowest mean. All methods receive two scalar rewards per round.}
\label{tab:structured-continuum}
\renewcommand{\arraystretch}{1.15}
\begin{tabular*}{\linewidth}{@{\extracolsep{\fill}}llrr@{}}
\toprule
\textbf{Learner} & \textbf{Model updates} & \textbf{Pure start} & \textbf{Mixed start}\\
\midrule
Checked commitment & 34-round prefix, then blocks & \RevCertZero & \RevCertMixed\\
Matched sparse NLS & Same checkpoints & \RevSparseZero & \RevSparseMixed\\
Dense NLS & Every round & \textbf{\RevDenseZero} & \textbf{\RevDenseMixed}\\
Oscillator posterior & Draw every round & \RevTSZero & \RevTSMixed\\
\bottomrule
\end{tabular*}
\end{table}

The posterior has a uniform prior over 262,144 full-circle frequencies and an isotropic Gaussian prior on the initial state. It integrates the state to update each frequency's selected-data likelihood, then samples a frequency and its conditional state. Disjoint development data select the prior scale; the appendix reports the three-resolution sensitivity check.

\begin{figure}[!ht]
\centering
\includegraphics[width=\linewidth]{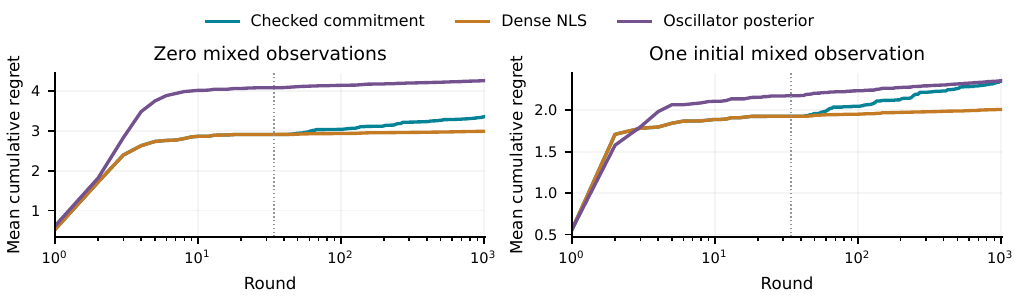}
\caption{Mean cumulative regret for the two initial sensing rules. The dotted line marks round 34, when the checked controller begins proposing longer blocks.}
\label{fig:global-commitment}
\end{figure}

With a pure start, checked commitment has 21.2\% lower mean regret than the posterior. The paired reduction is 0.90, with bootstrap interval $[0.46,1.36]$ adjusted for 95\% family coverage across the two posterior comparisons. One initial mixed observation closes this gap. Figure~\ref{fig:global-commitment} shows where the trajectories separate: most reward loss occurs early, while the later curves measure the cost of maintaining a forecast as time advances.

The checked and sparse policies take identical actions in all 384 primary runs. The controller's 1920 accepted checks attach prospective regret bounds to those decisions. The next comparison varies the time between fits and measures the resulting computational cost.

\clearpage
\subsection{Choosing prediction duration}
\label{sec:exp-certificates}
The confirmation catalog keeps the twelve frequency, amplitude, and noise combinations, with first-observation phases evenly spread around the cycle. Eight new seeds per cell give 96 paired instances. Every method uses a mixed start. For successive checkpoint times $n_k$, we compare doubling with the shorter rule $n_{k+1}=\lceil1.25n_k\rceil$, keeping both declared schedules and all matched controls.

\begin{table}[!ht]
\centering\small
\caption{Fresh confirmation at $T=1024$. Regret and refits use all 96 instances. Runtime is the median elapsed time on twelve prespecified first-seed instances, including confidence computation. The posterior updates a distribution rather than an NLS fit.}
\label{tab:confirmation-computation}\label{tab:refits}
\renewcommand{\arraystretch}{1.13}
\begin{tabular*}{\linewidth}{@{\extracolsep{\fill}}lrrr@{}}
\toprule
\textbf{Learner and checkpoint rule} & \textbf{Mean regret} & \textbf{Refits} & \textbf{Seconds}\\
\midrule
Checked commitment, factor 1.25 & 3.295 & 49 & 1.515\\
Checked commitment, doubling & 3.547 & 38 & 0.588\\
Matched sparse NLS, factor 1.25 & 3.295 & 49 & 0.227\\
Dense NLS & \textbf{3.246} & 1023 & 7.015\\
Oscillator posterior & 3.877 & -- & 5.441\\
\bottomrule
\end{tabular*}
\end{table}

The shorter checked schedule incurs 1.51\% more mean regret than dense fitting, with 95.2\% fewer nonlinear refits. The paired regret difference is 0.049 (95\% interval $[0.024,0.072]$). Its mean regret is 15.0\% below the posterior, with paired reduction 0.58 and adjusted interval $[0.06,1.11]$. The matched sparse control takes the same actions in all 96 instances; the runtime difference measures the additional confidence calculation.

\paragraph{Measure runtime.}
The median paired dense-to-checked runtime ratio is $4.69$ for the shorter schedule. Each ratio compares the same timed instance. Measurements run sequentially with one numerical-library thread after compilation and warmup. The common clock includes environment generation, decisions, and an identical compressed output payload. Appendix~\ref{app:runtime} gives all thirteen method and sensing combinations, both clock definitions, and the complete timing protocol.

\begin{figure}[!ht]
\centering
\includegraphics[width=\linewidth]{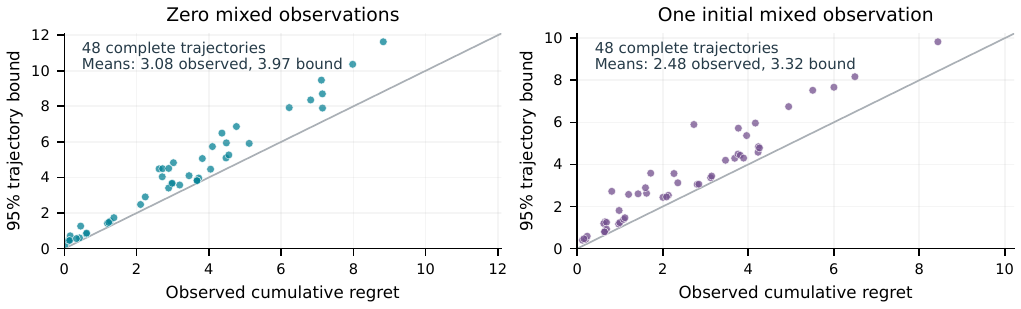}
\caption{Complete primary trajectories: measured regret against its confidence bound. The diagonal marks equality. Every point uses its own time-uniform confidence set and includes startup.}
\label{fig:trajectory-certificates}
\end{figure}

\paragraph{Certify the accumulated loss.}
A terminal confidence set bounds the regret of the entire selected action sequence. We compute 96 primary certificates (Figure~\ref{fig:trajectory-certificates}) from the first four prescribed seeds in each cell and sensing rule, plus 24 confirmation certificates from the first two seeds per cell. Each bound has 95\% coverage for its own trajectory and covers the full frequency circle and amplitude disk, including zero amplitude.

For the shorter-block confirmation subset, observed mean regret is 3.27 and the mean bound is 3.96. These certificates turn the collected rewards into a quantitative assessment of the executed run, complementing the prospective checks that selected its block lengths.

\clearpage
\subsection{Allocating observations between two oscillators}
\label{sec:exp-twoplane-new}
The two-plane experiment uses frequency pairs $(0.73,0.76)$ and $(0.73,1.17)$, amplitudes one, phases $(0,0.7)$, noise $0.25$, and horizon 8192. Eight paired seeds per cell give sixteen instances. Four dedicated startup rounds read both coordinates of each plane twice. Reward-seeking selections then supply the observations used for learning through round 256. The matched NLS control uses the same fits and checkpoints.

\paragraph{When is further sensing worth its cost?}
At each checkpoint, the controller compares the proposed decisions with a public schedule that spends another 256 rounds sensing, shortened at the horizon. At the first checkpoint, every run certifies an expected reward advantage of at least 190.01 for its proposed schedule. The information check also certifies that the planned observations measure both planes.

All 80 primary checks accept. The checked and matched NLS policies take the same actions and both have mean regret \RevTwoCert{}. Their complete-trajectory bounds range from 47.06 to 65.40, with the failure budget shared across the sixteen runs for simultaneous 95\% coverage. Appendix~\ref{app:exp} shows the sensing-cost curves; the supplied two-plane protocol regenerates the individual certificates.

\begin{figure}[!ht]
\centering
\includegraphics[width=\linewidth]{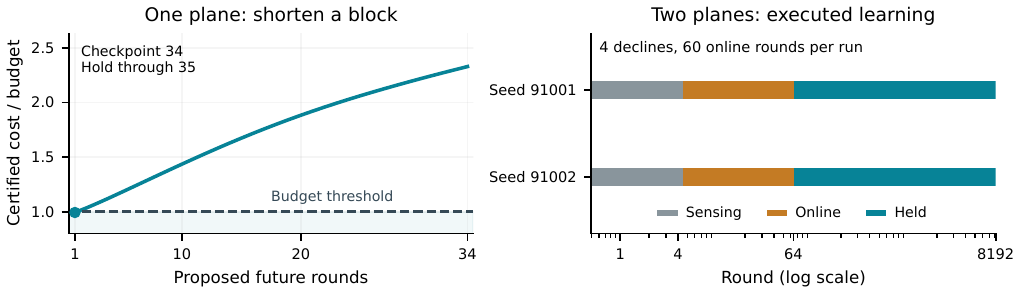}
\caption{The confidence check controls prediction duration. Left: on the first prescribed one-plane stress instance, only the one-round prefix of the block proposed at checkpoint 34 fits the budget. Right: both early-check two-plane runs decline four blocks, execute 60 online-learning rounds, and first hold a model after round 64.}
\label{fig:decision-checks}
\end{figure}

\paragraph{Let the check change the schedule.}
Two additional runs begin checking immediately after the four dedicated sensing rounds. The controller first gathers observations with its active policy. Final regrets are 74.23 and 67.29, below trajectory bounds 119.01 and 109.37. Figure~\ref{fig:decision-checks} shows the resulting learning intervals and a shortened one-plane block.

\paragraph{Exact and near resonances.}
The one-plane stress catalog crosses four frequencies with two amplitudes.
Against matched unchecked sparse NLS, checking reduces mean regret from
\ReviewStressPreviousSparseMean{} to \ReviewStressPreviousCheckedMean{}
on 32 previously inspected instances and from \ReviewStressHoldoutSparseMean{}
to \ReviewStressHoldoutCheckedMean{} on sixteen original holdout instances.
On the latter, the maximum observed regret falls from
\ReviewStressHoldoutSparseMaximum{} to \ReviewStressHoldoutCheckedMaximum{};
dense NLS has mean \ReviewStressHoldoutDenseMean{}.
On these instances, checked commitment uses \ReviewStressHoldoutCheckedFits{}
mean refits, versus \ReviewStressHoldoutSparseFits{} for sparse and
\ReviewStressHoldoutDenseFits{} for dense NLS. Across all 48 checked runs,
\ReviewStressAllLearningRounds{} learning rounds and
\ReviewStressAllShortenedBlocks{} shortened blocks adapt the prediction duration.
Appendix~\ref{app:stress-fresh} reports every cell and isolated runtime and memory costs.

\section{Conclusion}
\label{sec:discussion}\label{sec:freeze}
Selecting a measurement and deciding when to update its predictive model are coupled decisions. The relevant computational question is how long the current forecast remains useful for directing observations. A checked commitment answers this question using the data already collected. This makes model updates responsive to decision uncertainty and exposes their cost as part of the learning problem. The resulting separation between fitting, checking, and acting is useful when repeated nonlinear fits dominate computation. Extending this decision rule to many competing signals would require controlling which evolving states can remain unobserved and for how long.

\label{main-text-end}
\clearpage
\subsection*{AI use statement}
We used generative AI to help edit
prose, check internal consistency, design conceptual diagrams and implement diagnostic code. We reviewed the manuscript and take responsibility for
its text, claims, proofs, and figures.

\subsection*{Ethics statement}
The experiments use synthetic oscillator environments and include no newly collected human-subject data. The decision problem is a mathematical research model.

\subsection*{Reproducibility statement}
The supplementary package contains controller and confidence-oracle implementations, proof sources, frozen protocols, dependencies, result summaries, source hashes, independent audit records, and commands to regenerate the synthetic observations and rerun the comparisons. Appendix~\ref{app:exp} specifies the environments, seeds, evaluation cohorts, numerical settings, and timing protocol. The scripts generate potential noise arrays, selected-feedback traces, fit histories, and confidence outputs. Development and evaluation streams use distinct seeds. An explicit reproduction copy preserves the numerical source version bound by the frozen protocols.

\providecommand{\url}[1]{\texttt{#1}}

\clearpage
\appendix
\section{Proof guide and the core decision results}
\label{app:core}
The alias example and phase lower bound are proved first.
Section~\ref{app:committed-prediction} gives the complete proof of
Theorem~\ref{thm:committed-prediction}, including finite identification,
prediction transfer, and the removal of amplitude floors.
Section~\ref{struct:app:twoplane} proves the two-plane information lemma.
Appendix~\ref{app:active-rate} establishes the executed controller's
full-circle regret guarantee. The remaining appendices give confidence
coverage, trajectory and prospective bounds, numerical implementation,
and experimental protocols.

In the confidence and support derivations, $A_1$ denotes the supplied
upper amplitude bound $A_{\max}$.

\begin{proof}[Proof of Proposition~\ref{prop:alias-main}]

Both candidates predict $\cos(2\pi t/3)$ horizontally. Their vertical
predictions have opposite signs. At times congruent to $1,2,0$ modulo
three, the specified switching rule chooses horizontal, horizontal,
horizontal. Only the first of these choices is suboptimal, by $1+\sqrt3$.
The held reflection instead chooses vertical in round two, where its
prediction $\sqrt3/2$ differs from the observed $-\sqrt3/2$. Its regret
in the first two rounds is $(1+\sqrt3)+(\sqrt3-1)=2\sqrt3$; subsequent
regret is zero. For $T=1$, its regret is $1+\sqrt3<2\sqrt3$.
\end{proof}
\begin{proof}[Proof of Proposition~\ref{prop:phasefloor}]

Fix a policy $\pi$, and let $P_\pm^\pi$ be its action-and-observation transcript laws under the two instances. The claim is immediate when $\sigma=0$. Otherwise set
$\alpha=\min\{1/4,\sigma/(8A\sqrt T)\}$ and
$z_\pm=A(\cos\alpha,\pm\sin\alpha)$. Under either instance the
frequency is $\pi/4$. Every coordinate mean differs by at most
$2A\sin\alpha\le2A\alpha$ between the instances. The adaptive Gaussian
chain rule, with exactly two observed arms each round, gives
\[
 \operatorname{KL}(P_+^\pi,P_-^\pi)
 \le {4A^2T\alpha^2\over\sigma^2}\le {1\over16}.
\]
At each $t=1+4k\le T$, the preferred axes differ between the instances,
and each has coordinate gap $g=\sqrt2 A\sin\alpha$. Every feasible
pair $S$, including a mixed pair, has one-round pseudo-regrets $r_\pm(t,S)$ satisfying $r_+(t,S)+r_-(t,S)=2g$. Couple the two histories before selecting $S$.
Pinsker's inequality implies that their disagreement probability is at
most $1/\sqrt{32}$. Nonnegativity of regret then gives
\[
 \max_\theta\mathbb E_\theta R_T
 \ge \lceil T/4\rceil g(1-1/\sqrt{32})
 \ge {\sqrt2\over8}(1-1/\sqrt{32})AT\alpha.
\]
Here $\sin\alpha\ge\alpha/2$ and $\lceil T/4\rceil\ge T/4$.
Substituting the two possible values of $\alpha$ gives coefficients
larger than $1/64$ on both $AT$ and $\sigma\sqrt T$.
\end{proof}

\subsection{Proof of committed prediction}
\label{app:committed-prediction}
All times in this subsection are integers. Write
$\lambda(\alpha)=1-\max\{|\cos\alpha|,|\sin\alpha|\}$.
Let $d_{\mathbb T}$ denote the shortest angular distance modulo $2\pi$;
distance to a set is the minimum of those distances.
For a held candidate $(\nu,y)$, retain one selected coordinate $b_t$ per
round. It maximizes the candidate's coordinates, including when a mixed
pair is selected at a tie. Set
\[
 d_n=\sum_{t=1}^n
 \bigl[e_{b_t}^{\top}\{R(t\omega)z-R(t\nu)y\}\bigr]^2.
\]
The physical two-read discrepancy satisfies $D_n\ge d_n$. We prove the
geometric bound for $d_n$, uniformly over all greedy tie choices.

\paragraph{Finite identification under greedy measurements.}
Fix $0<A_0\le\|z\|\le A_1$, $\|y\|\le A_1$, and
$\lambda(\omega),\lambda(\nu)\ge\lambda_0>0$. Put
\[
 \begin{gathered}
 a_0=\arccos(1-\lambda_0),\qquad
 q_0=\left\lceil\max\{128,4\pi/a_0\}\right\rceil,\\
 m=\max\left\{q_0,\left\lceil\sqrt{1+12/\lambda_0}\right\rceil\right\},
 \qquad L=2m .
 \end{gathered}
\]
Any two consecutive coordinate rows at frequency $\alpha$ have Gram
minimum at least $\lambda(\alpha)$: their absolute inner product is either
$|\cos\alpha|$ or $|\sin\alpha|$. Thus $y=0$ cannot fit two observations
of the nonzero truth.

For nonzero states, identify $z,y$ with complex numbers $a,b$. Agreement on
one coordinate makes
\[
 p_t=\operatorname{Im}\!\left[(ae^{it\omega}-be^{it\nu})^2\right]=0.
\]
This sequence satisfies the degree-six recurrence with roots
$e^{\pm2i\omega},e^{\pm2i\nu},e^{\pm i(\omega+\nu)}$.
Six consecutive zeros therefore force $p_t=0$ for all subsequent times,
also when roots coincide. A real exponential polynomial has conjugate
coefficients at opposite frequencies. If $2\omega,2\nu,\omega+\nu$ are
distinct, their three-element support must contain a self-opposite
frequency. Nonresonance excludes $2\omega,2\nu\in\{0,\pi\}$, so
$\nu\in\{-\omega,\pi-\omega\}$ and $b=\pm\bar a$.
If frequencies coincide, either $\nu=\omega$ and $a=b$, or
$\nu=\omega+\pi$. In the latter case the merged coefficient cannot vanish,
since that would leave one non-self-opposite exponential. The remaining
two frequencies must be opposite, forcing $4\omega=\pi$ modulo $2\pi$.
These conclusions classify coordinate agreement, without presuming
future greedy agreement.

We next exclude the wrong classes within $L$ observed rounds.
If $d_{\mathbb T}(\alpha,\pi\mathbb Z)\ge a>0$ and
$q\ge\max\{128,4\pi/a\}$ is an integer, its first $q$ iterates, from any
starting phase, meet every open arc of length $17\pi/24$.
Indeed, Dirichlet approximation gives a reduced fraction $p/k$ with
$1\le k\le q$ and $|\alpha/(2\pi)-p/k|\le1/(kq)$.
The cases $k=1,2$ contradict the distance bound.
For $k\ge3$, the first $k$ points lie within $2\pi/q$ of an equally
spaced orbit, whose resulting largest gap is at most
$2\pi/3+4\pi/q\le2\pi/3+\pi/32<17\pi/24$.

Set $\epsilon=\pi/48$. For a true point $(u,v)$ of amplitude $A$, the arc
$(\pi+\epsilon,7\pi/4-\epsilon)$ has
$u+v<-\sqrt2A\sin\epsilon$ and $|v|>A\sin\epsilon$.
Its reflection $(u,-v)$ uniquely prefers coordinate 2 and disagrees
there. On $(3\pi/4+\epsilon,3\pi/2-\epsilon)$ the reflection
$(-u,v)$ uniquely prefers coordinate 1 and disagrees there.
Both arcs have length $17\pi/24$.
Since the distances of $\omega$ and $2\omega$ from $\pi\mathbb Z$ are
at least $a_0$ and $2a_0$, respectively, ordinary reflections are
contradicted by time $q_0$, and even-time reflections by $2q_0$.
For the exceptional half-turn family, suppose first $a\ne b$.
At even times the nonzero difference is axis-aligned, so agreement
requires selection of its zero coordinate. Comparing opposite candidate
states at times 2 and 6 forces a candidate tie at time 2; times 4 and 8
force another at time 4. Those states differ by an odd quarter turn,
which is impossible for a nonzero state. If $a=b$, candidate and truth
are opposite at odd times. Agreement either is already impossible at
time 1, or forces its zero coordinate; the opposite candidate at time 5
cannot prefer that zero coordinate as well. Hence
\[
 d_L=0\quad\Longrightarrow\quad \nu=\omega\pmod{2\pi},\quad y=z.
 \tag{G1}\label{eq:audit-finite-identification}
\]

\begin{proposition}[Prediction from a positive-amplitude reference]
\label{struct:prop:committed-obs}
Under the bounds above, there are constants $c,c_1,K_0>0$, depending
only on $(A_0,A_1,\lambda_0)$, such that, with
$\Delta=d_{\mathbb T}(\omega,\nu)$,
\[
 \begin{aligned}
 d_L&\ge c(\|z-y\|^2+\Delta^2),\\
 d_n&\ge c_1A_0^2n\min\{n^2\Delta^2,1\},\\
 \max_{t\in\{0,\ldots,2n\}}\|R(t\omega)z-R(t\nu)y\|^2
 &\le K_0d_n/n
 \end{aligned}
\]
for every $n\ge2L$ in the last two inequalities.
\end{proposition}
\begin{proof}
Let $J=R(\pi/2)$ and $r_t=R(-t\omega)e_{b_t}$.
For an initial-state perturbation $h$ and frequency perturbation $\gamma$,
put $v=\gamma Jz$ and $\xi=h+m v$. Pairing rounds $2r-1,2r$ and using
$(u+w)^2\ge u^2/2-w^2$ gives
\[
 \sum_{t=1}^{2m}[r_t^\top(h+tv)]^2
 \ge \frac{\lambda_0m}{2}\|\xi\|^2+
 m\left\{\frac{\lambda_0(m^2-1)}6-1\right\}\|v\|^2
 \ge\frac{\lambda_0m}{16}
       \{\|\xi\|^2+m^2A^2\gamma^2\},
\]
where $A=\|z\|$; the odd times have mean $m$ and centered squared sum
$m(m^2-1)/3$. Since
\[
 \|h\|^2+\gamma^2\le
 [2+(mA_0)^{-2}]\{\|\xi\|^2+m^2A^2\gamma^2\},
 \qquad
 s_0=\frac{\lambda_0m}{16[2+(mA_0)^{-2}]},
\]
the selected Jacobian has squared minimum singular value at least $s_0$.
The rotation inequalities
$\|R(u)-I\|\le|u|$ and $\|R(u)-I-uJ\|\le u^2/2$
bound its nonlinear remainder by
$K_{\rm rem}(\|h\|^2+\gamma^2)$, where
\[
 K_{\rm rem}=\tfrac12\sqrt L(L+A_1L^2),\qquad
 \rho=\min\{1,\sqrt{s_0}/(2K_{\rm rem})\}.
\]
Thus $d_L\ge s_0(\|z-y\|^2+\Delta^2)/4$ inside radius $\rho$.

Outside a local neighborhood, let $M_0$ be the minimum of the selected
greedy discrepancy over the compact class where
$\|z-y\|^2+|e^{i\omega}-e^{i\nu}|^2\ge\rho^2/4$.
At each tie minimize over both eligible coordinates. This objective is
lower semicontinuous, because the minimizing coordinate set expands at
a tie. Equation~\eqref{eq:audit-finite-identification} excludes every
zero, so $M_0>0$. Circular distance is at most $\pi/2$ times chordal
distance. The local and complementary bounds therefore imply the first
claim with the explicit choice
\[
 c=\tfrac14\min\{s_0/4,M_0/(4A_1^2+4)\}.
\]
This defines the constant by a finite compact minimization.

For $n\ge2L$, let $M=\lfloor n/L\rfloor\ge2$.
Apply the first inequality at each complete block origin, rotating the
two initial states separately. In complex notation,
\[
 d_n\ge c\sum_{j=0}^{M-1}|ae^{ijL\omega}-be^{ijL\nu}|^2+cM\Delta^2.
 \tag{G2}\label{eq:audit-block-aggregation}
\]
If $\Delta\ge\pi/(4L)$, the second term is at least
$c\pi^2n/(32L^3)$. Otherwise choose local frequency representatives with $|\omega-\nu|=\Delta$, and put $\iota=L(\omega-\nu)$, so $|\iota|<\pi/4$.
Profiling the first sum over unrestricted $b$ yields
\[
 A^2\left(M-\frac{|S_M|^2}{M}\right),\quad
 S_M=\sum_{j=0}^{M-1}e^{ij\iota},\quad
 M-\frac{|S_M|^2}{M}
 =\frac2M\sum_{r=1}^{M-1}(M-r)(1-\cos r\iota).
\]
When $M|\iota|\le4$, restrict to $r\le\lfloor M/2\rfloor$ and use
$1-\cos(r\iota)\ge r^2\iota^2/4$ to obtain
$M^3\iota^2/324$. When $M|\iota|>4$, the geometric-sum bound
$|S_M|\le\pi/|\iota|$ gives a value above $M/4$.
Consequently it is at least
$(M/512)\min\{M^2\iota^2,1\}$.
Using $ML\ge n/2$ and $A\ge A_0$ proves the second claim with
\[
 c_1=\min\{c/(4096L),c\pi^2/(32A_0^2L^3)\}.
\]
The separate $cM\Delta^2$ term in
\eqref{eq:audit-block-aggregation} controls aliases of the block stride.

For any integer $0\le t\le2n$, comparison with each block origin gives
\[
 \|R(t\omega)z-R(t\nu)y\|^2
 \le \frac{2d_n}{cM}+8A_1^2n^2\Delta^2.
\]
If $n\Delta\le1$, the second claim bounds
$n^2\Delta^2$ by $d_n/(c_1A_0^2n)$.
Otherwise it gives $d_n\ge c_1A_0^2n$, while the state error is at most
$4A_1^2$. Both cases imply the prediction inequality with
$K_0=4L/c+8A_1^2/(c_1A_0^2)$.
\end{proof}

\begin{proof}[Proof of Theorem~\ref{thm:committed-prediction}]
Put $a=\max\{\|z\|,\|y\|\}$ and $\kappa=\sqrt{\lambda_0/12}$.
If $a=0$, all means and regrets vanish. Otherwise normalize both states
by $a$, which preserves every greedy tie and scales squared errors by
$a^{-2}$. If both normalized amplitudes are at least $\kappa$, apply
Proposition~\ref{struct:prop:committed-obs} with $A_0=\kappa,A_1=1$.
Its threshold $2L$ and constant $K_0$ depend only on $\lambda_0$.
Scaling back proves the same prediction inequality for the original states.

Otherwise, let $v_t,u_t$ denote the retained values of the original,
unnormalized larger and smaller signals.
The adjacent-row Gram at the larger signal's frequency
and the amplitude bound on the smaller give
\[
 \begin{gathered}
 \sum v_t^2\ge\lambda_0\lfloor n/2\rfloor a^2,\qquad
 \sum u_t^2\le n\kappa^2a^2,\\
 D_n\ge d_n\ge
 \left(\frac{\lambda_0}{2}\lfloor n/2\rfloor-n\kappa^2\right)a^2
 \ge\frac{\lambda_0}{12}na^2
 \end{gathered}
\]
for $n\ge2$, using $(v-u)^2\ge v^2/2-u^2$.
The full-state squared error is at most $4a^2$ at every time.
Thus $H=2L$ and $K=\max\{K_0,48/\lambda_0\}$ work in all cases.

Finally, write $\Delta_i(t)=\mu_i(t)-\mu_i^{(\nu,y)}(t)$.
For a true optimal pair $A_t^*$ and the candidate's greedy pair $S_t$,
\[
 \sum_{i\in A_t^*}\mu_i(t)-\sum_{i\in S_t}\mu_i(t)
 \le\sum_{i\in A_t^*}\Delta_i(t)-\sum_{i\in S_t}\Delta_i(t)
 \le4\max_i|\Delta_i(t)|.
\]
Each arm error is bounded by the state norm error. Summation gives
$R_{1:n}\le4\sqrt{KnD_n}$. Rotating each initial state to a shifted block
origin preserves the argument.
\end{proof}

\subsection{Information from two-plane greedy decisions}
\label{app:spatial-geometry}
\label{struct:app:twoplane}
\label{struct:sec:restricted-twoplane}
Use the four distinct arms of Section~\ref{sec:twoplane}, with
$A^{(j)}=\|x_j\|>0$. A block of $\ell$ rounds has phase discrepancy at
most $1/48$ in plane $j$ if, for every circular arc $I$,
\[
 \left|\frac{\#\{t\text{ in block}:t\omega_j\bmod2\pi\in I\}}{\ell}
             -\frac{|I|}{2\pi}\right|\le\frac1{48}.
\]
The arc convention includes endpoint limits. This condition is invariant
under phase shifts; requiring it for $\ell$ consecutive phases therefore
covers every shifted block. Set
$\varepsilon_*=\sin(\pi/48)$,
$\eta_{\rm sp}=\sin^2(\pi/16)/12$, and
$d_{j,t,b}=R(-t\omega_j)e_b$.

\begin{lemma}[Information from accurate greedy decisions]
\label{struct:lem:spatial}
Let $\widehat\mu_{j,b}(t)$ denote predicted means. Suppose both planes have phase discrepancy at most $1/48$ on an $\ell$-round block. If
$|\widehat\mu_{j,b}(t)-\mu_{j,b}(t)|\le\varepsilon_*A^{(j)}$
for all four arms throughout that block, then selecting the two largest
predicted means, with any tie rule, gives
\[
 \sum_{t\text{ in block}}\sum_{b:(j,b)\in S_t}
 d_{j,t,b}d_{j,t,b}^{\top}\succeq\eta_{\rm sp}\ell I_2,
 \qquad j=1,2.
\]
\end{lemma}
\begin{proof}
Fix a plane $j$, the other plane $k$, and a nonzero vector $h$.
If some prediction of $j$ is positive and some prediction of $k$ is
negative, a top-two pair includes a $j$-arm. Failure of this sufficient
condition lies in the union of two bad-sign events: both true coordinates
of $j$ are at most $\varepsilon_*A^{(j)}$, or both coordinates of $k$
are at least $-\varepsilon_*A^{(k)}$.
Each event is a circular arc of length
$\pi/2+2\arcsin(\varepsilon_*)$.

Put $a=\sin(\pi/16)$. For either of the two possible coordinate rows of
$j$, phases with $|d_{j,t,b}^{\top}h|<a\|h\|$ occupy two arcs.
Their union over both coordinates uses four arcs of total normalized
length $4\arcsin(a)/\pi=1/4$.
Applying the two marginal discrepancy bounds to these six arcs leaves
at least the fraction
\[
 1-2\left(\frac14+\frac{\arcsin\varepsilon_*}{\pi}\right)
   -\frac{4\arcsin a}{\pi}-\frac6{48}
 =\frac1{12}
\]
of the rounds. On each, a $j$-arm is selected and either possible row
has squared projection at least $a^2\|h\|^2$.
Thus the Gram quadratic form is at least
$a^2\ell\|h\|^2/12$. Since $h$ was arbitrary, the matrix inequality follows.
No joint phase assumption or frequency separation is used.
\end{proof}

\section{An active regression policy on learning rounds}
\label{app:active-rate}
\subsection{An active fallback with explicit finite-horizon constants}
\label{app:active-rate-recovery}
This construction covers every frequency on the circle and every initial
state in the disk $\|z\|_2\le A$, where $A=A_{\max}>0$ is supplied. It uses the
same selected feedback as the controller. The one-plane model has two
replica pairs; the two-plane model has four distinct coordinate arms and
all six pairs are available. The physical arm-noise vector is conditionally centered and jointly
sub-Gaussian with supplied variance proxy $\sigma^2 I$. The horizon $T$ is supplied.

\paragraph{The public controller.}
Fix a predictable prefix of $n_0\le T$ rounds and increasing checkpoints.
At a checkpoint $n$, choose an endpoint $m>n$ and a proposed pair schedule
from the observed history. Let $\overline U_{n,m}$ be an outward upper bound on its
regret maximized over a simultaneous confidence set. The confidence
construction receives failure budget $\delta/2$. For $\sigma=0$, use the
zero selected-loss set $\{\theta:L_n(\theta)=0\}$; the true model belongs
to it deterministically. Accept the proposed
schedule precisely when
\begin{equation}
 \overline U_{n,m}\le c\sigma\sqrt{\log(8T/\delta)}(\sqrt m-\sqrt n).
 \label{eq:recovery-gate}
\end{equation}
A failed or unresolved check activates the learner below during this
block. It receives only the observations on its own active rounds, using
the original physical time as its context. Extra checks may reject a
block but may never override a failed test in~\eqref{eq:recovery-gate}.
This defines an action in every round without waiting for a later
initialization deadline.

\paragraph{The online oscillator oracle.}
Put $d=\delta/2$. For one plane let $(J,K,v,h)=(1,2,\sigma^2/2,1)$;
for two planes let $(J,K,v,h)=(4,6,\sigma^2,2)$. Here $J$ counts scalar
regression banks, $K$ counts physical pair choices, $v$ is one bank's
noise variance proxy, and $h$ is the number of bank observations per round.
Define
\begin{align}
 B&=A+\sqrt{2v\log(8hT/d)}, &
 S_T&=\sum_{t=1}^T t^2,\notag\\
 M&=\max\{2,\lceil\pi\sqrt{S_T}\rceil\}, &
 D&=(\pi A/M)^2S_T\le A^2,\notag\\
 a&=2B^2/A^2, &
 Q&=aA^2+2B^2\log(1+T/(2a))+2B^2\log M.
 \label{eq:recovery-oracle-constants}
\end{align}
The frequencies $2\pi j/M$, $0\le j<M$, cover the entire circle.
Each bank has an independent two-dimensional Vovk--Azoury--Warmuth
regressor at each frequency \citep{vovk1998regression,azoury2001relative}. For coordinate $\beta\in\{1,2\}$ at physical time $t$ and grid frequency $\nu$, use the sensing column $x_t=R(-t\nu)e_\beta$. Let $y_s$ denote earlier labels in that bank, clipped to $[-B,B]$. Set $V=aI+\sum_{s<t}x_sx_s^\top$ and $b=\sum_{s<t}y_sx_s$ within the bank. Its expert prediction is
\[
 u_t=\operatorname{clip}_{[-B,B]}
 \left({x_t^\top V^{-1}b\over1+x_t^\top V^{-1}x_t}\right).
\]
Combine the $M$ experts
by the square-loss aggregating algorithm \citep{vovk2001statistics} with $\eta=1/(2B^2)$ and
initial weights $1/M$. If their current predictions are $u_j$ and
normalized weights are $w_j$, the master prediction is
\[
 \widehat y={\log\sum_jw_j e^{-\eta(B-u_j)^2}
                 -\log\sum_jw_j e^{-\eta(-B-u_j)^2}\over4B\eta}.
\]
For the current label $y$ clipped to $[-B,B]$, update $w_j$ proportionally to $w_j e^{-\eta(y-u_j)^2}$.

For one plane, the two candidate predictions are the master predictions
for horizontal and vertical; the observed label is the average of the
two selected replica rewards. For two planes, index four independent
banks by $(s,j)\in\{1,2\}\times\{1,2\}$, where $s$ is the position of
an arm in the selected pair sorted by arm index and $j$ is its plane.
The prediction for pair $(i_1,i_2)$ is the average of the corresponding
two bank predictions. Both predictions are made before either reward
is revealed. Update each selected bank with its one coordinate reward.
This factorization uses $4M$ two-dimensional regressors, without a
Cartesian product of frequencies.

\paragraph{Active action selection.}
Let $f_t(a)$ denote these predicted coordinate or pair-average rewards.
For an exploration parameter $\gamma>0$, with $b_t$ the first maximizer under a fixed tie order, use
\[
 p_t(a)={1\over K+\gamma(f_t(b_t)-f_t(a))}\quad(a\ne b_t),
 \qquad p_t(b_t)=1-\sum_{a\ne b_t}p_t(a).
\]
Sample a pair using independent controller randomness. This is the
reward form of SquareCB \citep{foster2020beyond}. Its one-step inequality
applies after scaling from a bounded interval and therefore also to the
unnormalized reward units here. Only active rounds update its banks.

\paragraph{Explicit coefficient.}
For any public $0<\rho<1/2$ and $k>0$, put
\begin{align}
 E_b&={Q+(1+2\rho)D+2v\log(4J/d)/\rho\over1-2\rho}, &
 E&={J E_b\over h},\notag\\
 C&={kE+(A+B)^2\log(4/d)\over1-e^{-k}}, &
 G&=2\sqrt2 A,\notag\\
 \gamma&=\sqrt{8KT/C}, &
 F_T&=2\sqrt{2KTC}+G\sqrt{T\log(4/d)/2}.
 \label{eq:recovery-finite-coefficient}
\end{align}
When $v=0$, take the limiting value $E_b=Q+D$. The implementation
minimizes this explicit bound using public parameters alone; every
positive $k$ and admissible $\rho$ gives a valid bound.

\begin{proof}[Proof of Theorem~\ref{thm:reconstructed-main}]
We give each of four fallback events failure budget $d/4$.
First, at most $hT$ bank observations occur. The sub-Gaussian tail bound and
the definition of $B$ imply that every unmodified label lies in
$[-B,B]$ except on an event of probability at most $d/4$. Thus clipping
agrees with the physical labels on this event.

For one expert, write $r=x^\top V^{-1}b$ and
$\ell=x^\top V^{-1}x$. The regularized least-squares potential increases
by $(y-r)^2/(1+\ell)$ after adding $(x,y)$. Before clipping,
\[
 \left(y-{r\over1+\ell}\right)^2
 -{(y-r)^2\over1+\ell}
 ={y^2\ell\over1+\ell}-{r^2\ell\over(1+\ell)^2}
 \le B^2\log(1+\ell).
\]
Clipping can only reduce loss for $|y|\le B$. Telescoping and the
matrix determinant lemma bound the expert's regret to any state $z$ by
$a\|z\|^2+B^2\log\det(I+\sum xx^\top/a)$. Unit rows and concavity of
$\log$ bound this by $aA^2+2B^2\log(1+T/(2a))$.
The square-loss aggregating algorithm adds $2B^2\log M$.
Consequently each bank has squared-loss regret at most $Q$ to every
state of norm at most $A$ at every grid frequency.

Choose the nearest circular grid frequency to the truth and keep the
true initial state. At physical time $t$ its coordinate error is at most
$A\pi t/M$. Thus its cumulative squared error on any selected subsequence
is at most $D$. Let $\mu_s$ be the true mean of bank observation $s$, $g_s$ its grid-comparator prediction, and $\varepsilon_s$ its noise. Write $e_s=\widehat y_s-\mu_s$ for the master's error and $b_s=g_s-\mu_s$ for the comparator's error. These quantities are
predictable, including the selected coordinate. Conditional sub-Gaussian
exponential supermartingales give, with failure probability $d/(4J)$,
\[
 2\sum_s\varepsilon_s(e_s-b_s)
 \le \rho\sum_s(e_s-b_s)^2
             +{2v\over\rho}\log(4J/d).
\]
On the label-envelope event, the deterministic regression guarantee
implies
\[
 \sum_s e_s^2\le Q+\sum_s b_s^2+
                    2\sum_s\varepsilon_s(e_s-b_s).
\]
Use $(e-b)^2\le2e^2+2b^2$ and $\sum b_s^2\le D$ to obtain
$\sum e_s^2\le E_b$. Union over banks costs $d/4$. For one plane this
bounds the realized chosen-action error by $E_b$; for two planes,
$((e_1+e_2)/2)^2\le(e_1^2+e_2^2)/2$ gives the bound $E=2E_b$.

Let $\mathcal F_{t-1}$ contain the history immediately before round $t$'s action draw, including whether it is active. Write $\bar\mu_t(a)$ for the true coordinate mean in one plane or pair-average mean in two planes. For the sampled active choice $a_t$, put $Z_t=(f_t(a_t)-\bar\mu_t(a_t))^2$; set $Z_t=0$ on inactive rounds. To bound their conditional means, use $0\le Z_t\le H=(A+B)^2$ and
\[
 \mathbb E[e^{-kZ_t/H}\mid\mathcal F_{t-1}]
 \le1-(1-e^{-k})\mathbb E[Z_t\mid\mathcal F_{t-1}]/H.
\]
An exponential supermartingale and Markov's inequality imply
\[
 \sum_t\mathbb E[Z_t\mid\mathcal F_{t-1}]
 \le {k\sum_tZ_t+H\log(4/d)\over1-e^{-k}}\le C
\]
except with probability $d/4$. Masking inactive rounds by zero preserves
predictability and this argument, even when activation depends on the
entire earlier history.

SquareCB's one-step inequality bounds coordinate or pair-average regret
by $2K/\gamma+(\gamma/4)\mathbb E[Z_t\mid\mathcal F_{t-1}]$.
Physical pair rewards are twice these rewards, so conditional physical
regret summed over active rounds is at most
$4KT/\gamma+\gamma C/2=2\sqrt{2KTC}$. Physical regret lies in $[0,G]$:
for one plane this follows from $|(e_1-e_2)^\top R(t\omega)z|\le\sqrt2 A$; for two planes each
plane contributes at most $\sqrt2 A$ to the difference between two pairs.
The conditional Hoeffding inequality adds
$G\sqrt{T\log(4/d)/2}$ with failure probability at most $d/4$.
This proves the active fallback bound $F_T$ with failure probability $d$.

On the confidence event every accepted block costs at most its $\overline U_{n,m}$.
Their disjoint increasing intervals telescope under
\eqref{eq:recovery-gate}; inserting inactive gaps can only increase the
sum to $\sqrt T-\sqrt{n_0}$. The prefix costs at most $Gn_0$.
Unioning the confidence and fallback events proves the theorem.
Finally $\log M=O(\log T)$, $B^2=O(A^2+\sigma^2\log(T/d))$, and one
may use any fixed $\rho\in(0,1/2)$ and $k>0$, giving
$F_T=\widetilde O(\sqrt T)$.
\end{proof}

\section{Continuous confidence and trajectory bounds}
\subsection{An anytime confidence set from selected feedback}
\label{sec:gaussian-mixture-confidence}
Let $\theta_*$ denote the true parameter. Let $d_i(\omega)$ be the column sensing vector whose transpose is
observation row $i$ of the planar design at time $t_i$. The selected pair at each round is chosen before its noise
is revealed. Through round $n$, write $X_\omega$ for the matrix of these
rows, $y$ for their rewards, and
$L_n(\omega,z)=\|y-X_\omega z\|_2^2$ for their squared loss.
Each physical replica observation contributes a separate row.

Let $\epsilon_t$ collect the selected noises, and let $\mathcal G_t$ contain
the past and the selected pair, before its rewards are revealed. Conditional
joint $\sigma$-sub-Gaussian noise means
\[
 \mathbb E[\exp(u^\top\epsilon_t)\mid\mathcal G_t]
 \le\exp(\sigma^2\|u\|_2^2/2),\qquad u\in\mathbb R^{|S_t|}.
\]
For $\sigma>0$, fix $s_{\rm prior}>0$ before collecting data and use the independent
prior $\omega\sim\operatorname{Unif}[0,2\pi)$, $z\sim N(0,s_{\rm prior}^2I_2)$.
For a fixed alternative parameter, let $v_t$ be its $\mathcal G_t$-measurable
vector of mean differences from the true model. The bound above gives
\[
 E\!\left[\exp\!\left\{
 \epsilon_t^\top v_t/\sigma^2-\|v_t\|^2/(2\sigma^2)
 \right\}\,\middle|\,\mathcal G_t\right]\le1.
\]
Consequently the exponential squared-loss score ratio for that alternative
is a nonnegative supermartingale at complete round boundaries. Integrating
these ratios against the fixed prior preserves this property by Tonelli's
theorem. Ville's inequality therefore gives
one event of probability at least $1-\delta$ on which, at every time,
\begin{equation}\label{eq:gaussian-mixture-cutoff}
 L_n(\theta_*)\le C_n-2\sigma^2\log(\delta J_n),\qquad
 J_n=\frac1{2\pi}\int_{-\pi}^{\pi} e^{g_n(\omega)}\,d\omega,
\end{equation}
where $C_n=\|y\|_2^2$, $\alpha=\sigma^2/s_{\rm prior}^2$,
$G_\omega=X_\omega^\top X_\omega$, $h_\omega=X_\omega^\top y$, and
\[
 g_n(\omega)=-\tfrac12\log\det(I_2+G_\omega/\alpha)
 +\frac{h_\omega^\top(G_\omega+\alpha I_2)^{-1}h_\omega}{2\sigma^2}.
\]
The Gaussian state integral is an analytic identity for the squared-loss
score. The prior is a proof device. The confidence set may subsequently be
intersected with any supplied amplitude disk; the argument also covers
zero amplitude and resonant frequencies.

\paragraph{A validated lower integral.}
A proposed center frequency $c$ need not minimize any objective.
Normalizing the center's Gaussian state score gives a Gaussian distribution with
$\mu=(G_c+\alpha I_2)^{-1}h_c$ and
$V=\sigma^2(G_c+\alpha I_2)^{-1}$.
Let $r=y-X_c\mu$, and write $X'_c$ for the frequency derivative of $X_c$.
Define
\[
 a=r^\top X'_c\mu-\operatorname{tr}(X_c^\top X'_cV),\qquad
 B=\|\mu\|\sum_i t_i^2|r_i|
 +(\|\mu\|^2+2\operatorname{tr}V)\sum_i t_i^2.
\]
For every circular arc with radius $0<h\le\pi$,
\begin{equation}\label{eq:mixture-local-integral}
 \log J_n\ge g_n(c)+\log(h/\pi)
 -\frac{h|a|}{\sigma^2}-\frac{h^2B}{2\sigma^2}.
\end{equation}
To prove this, take the ratio of the state-integrated scores at $c+u$
to that at $c$.  Jensen's inequality lower-bounds its logarithm by
$-E[L_n(c+u,z)-L_n(c,z)\mid c,y]/(2\sigma^2)$.
The row identities $\|d_i\|=1$, $\|d_i'\|=|t_i|$, and
$\|d_i''\|=t_i^2$ give
\[
 E[L_n(c+u,z)-L_n(c,z)\mid c,y]\le 2|u||a|+u^2B.
\]
Indeed the mean residual contributes the linear term and a remainder at
most $u^2\|\mu\|\sum_i t_i^2|r_i|$; the squared mean displacement
contributes at most $u^2\|\mu\|^2\sum_i t_i^2$.
The second derivative of $d_i(c+u)^\top Vd_i(c+u)$ has absolute value at
most $4t_i^2\operatorname{tr}V$, giving the remaining covariance term.
Integrating this uniform lower bound over the arc proves
\eqref{eq:mixture-local-integral}.

The software evaluates the center quantities and the final expression
with outward arbitrary-precision interval arithmetic.  The numerical
optimizer proposes only $c$ and $h$.  A lower bound on $J_n$ enlarges the
cutoff in \eqref{eq:gaussian-mixture-cutoff}, preserving coverage.
The same fixed prior and failure probability apply at every adaptive
query; no union bound over calls or proposed centers is needed.

The trajectory supports and their polygon inflation are derived in
Appendix~\ref{rebuild:app:finite-certificate}; the arithmetic contract is
specified in Appendix~\ref{app:coefficient-contract}.

\subsection{Certificates for a complete observed trajectory}
\label{rebuild:app:finite-certificate}

A trajectory certificate uses the rewards already collected to bound the
regret of the actions actually taken. Its confidence statement concerns
the unknown mean model. It neither observes the unchosen rewards nor
requires the controller to have used the same confidence set online.

Fix a horizon $T$. A selected pair $S_t$ is chosen before observing its
round-$t$ rewards. Let $\mathcal F$ be the finite family of feasible pairs,
and let $\mu_A^\theta(t)$ denote the sum of a pair's mean rewards under
model $\theta$. Define the model-dependent trajectory loss in reward units
by
\[
 \mathcal R_T(\theta)=\sum_{t=1}^T
 \max_{A\in\mathcal F}\{\mu_A^\theta(t)-\mu_{S_t}^\theta(t)\}.
\]
The actual pseudo-regret is $R_T=\mathcal R_T(\theta_*)$.
Suppose a random confidence set $\mathcal C_T$ satisfies
$\Pr\{\theta_*\in\mathcal C_T\}\ge1-\delta$ under this predictable
policy. The fixed-prior mixture construction of
Section~\ref{sec:gaussian-mixture-confidence} supplies such a set from
selected observations. If its coverage is simultaneous in time, the
following result also applies at a data-dependent stopping time bounded
by the stated horizon.

\begin{theorem}[Whole-trajectory regret certificate]
\label{rebuild:thm:trajectory-certificate}
Any validated upper bound $\mathcal B_T$ on
$\sup_{\theta\in\mathcal C_T}\mathcal R_T(\theta)$ satisfies
\[
 \Pr\{R_T\le\mathcal B_T\}\ge1-\delta.
\]
If $\mathcal C_T\subseteq\bigcup_{c\in\mathcal K}\mathcal C_c$ is a
complete finite cover and $B_c$ bounds
$\sup_{\theta\in\mathcal C_c}\mathcal R_T(\theta)$, then
$\mathcal B_T=\max_c B_c$ is valid. A known deterministic regret bound can
be intersected with this certificate by taking the smaller value.
\end{theorem}
\begin{proof}
On the coverage event, the true model lies in one of the covering cells,
and its actual action sequence is precisely the sequence used in
$\mathcal R_T$. Each displayed upper bound therefore applies to
$\mathcal R_T(\theta_*)$. All rounds are evaluated inside the same
confidence event; no additional union bound over trajectory times is used.
\end{proof}

\paragraph{One oscillator and arbitrary selected pairs.}
Let $x\in\mathbb R^2$ be the initial state, $\nu$ its frequency on the
circle, and $\|x\|\le A_1$. No amplitude lower bound or nonresonance
condition is required here. For a pair $A$, let $v_A\in\mathbb R^2$ count
its horizontal and vertical reads. The feasible count vectors are
$(2,0)$, $(1,1)$, and $(0,2)$. Define
\[
 v_{t,A}=v_A-v_{S_t},\qquad
 a_{t,A}(\nu)=R(-t\nu)v_{t,A}.
\]
Then $\mu_A^{(\nu,x)}(t)-\mu_{S_t}^{(\nu,x)}(t)
=a_{t,A}(\nu)^\top x$. Including $A=S_t$ makes each per-round maximum
nonnegative. The deterministic bound is $R_T\le2\sqrt2A_1T$.

Consider a frequency cell centered at $c$ with circular radius $h\le\pi$.
Suppose every compatible state in that cell lies in a ball with center
$z_c$ and radius $r_c$. Set
\[
 e_{c,t,A}=A_1\|v_{t,A}\|\min\{2,t h\}.
\]
The rotation inequality
$\|R(t\nu)-R(tc)\|_{\rm op}\le\min\{2,t h\}$ gives
\begin{equation}
 \mathcal R_T(\nu,x)\le
 F_c(x):=\sum_{t=1}^T\max_{A\in\mathcal F}
 \{a_{t,A}(c)^\top x+e_{c,t,A}\}.
 \label{rebuild:eq:cell-trajectory-function}
\end{equation}
In particular the elementary support bound
\begin{equation}
 B_c^{\rm support}=\sum_{t=1}^T\max_{A\in\mathcal F}
 \min\left\{A_1\|v_{t,A}\|,
 a_{t,A}(c)^\top z_c+r_c\|v_{t,A}\|+e_{c,t,A}\right\}
 \label{rebuild:eq:cell-support-certificate}
\end{equation}
is valid. The first quantity inside the minimum is an independent physical
amplitude bound on the action difference. Taking the maximum over cells
\emph{after} summing over times preserves a common frequency cell for the
whole trajectory.

\paragraph{Retaining a common state across rounds.}
There is a tighter finite alternative to separate per-round ball supports.
For an integer $q\ge3$, the regular polygon with vertices
\[
 z_{c,k}=z_c+\frac{r_c}{\cos(\pi/q)}
       (\cos(2\pi k/q),\sin(2\pi k/q)),\qquad 0\le k<q,
\]
contains the entire state ball. Since $F_c$ in
\eqref{rebuild:eq:cell-trajectory-function} is convex,
\begin{equation}
 B_c^{\rm polygon}=\max_{0\le k<q} F_c(z_{c,k})
 \label{rebuild:eq:polygon-certificate}
\end{equation}
is a valid cell bound. Indeed every point of the ball is a convex
combination of polygon vertices, and convexity bounds its value by the
largest vertex value. The polygon may extend outside the amplitude disk;
the frequency remainder $e_{c,t,A}$ still uses $A_1$ because that remainder
is applied to the true compatible state before the convex extension.
The minimum of \eqref{rebuild:eq:cell-support-certificate},
\eqref{rebuild:eq:polygon-certificate}, and $2\sqrt2A_1T$ is also valid.
Outward arithmetic must enclose polygon vertices, support evaluations,
and the final maximum; an unquantified floating-point inflation is not a
certificate.
For example, if computed unit directions are within Euclidean distance
$e$ of their ideal values, their convex hull has support at least
$\cos(\pi/q)-e$ in every unit direction. Inflating the circumradius by
$1+8e$ therefore retains the state ball whenever
$8\cos(\pi/q)\ge1+8e$. A separately validated vertex-position error can
be included in the same support inequality, or in the evaluation error
of the trajectory functional.

\paragraph{Constructing complete state balls from selected losses.}
Index the selected scalar observations by $i$, with times $t_i$ and axes
$b_i$. Write $X(\nu)$ for the matrix with rows
$e_{b_i}^\top R(t_i\nu)$, and $y$ for their observed rewards. Each physical
replica contributes its own row. Suppose a validated confidence cutoff
$C\ge0$ defines
\[
 \mathcal C_T=\{(\nu,x):\|x\|\le A_1,
                            \|y-X(\nu)x\|^2\le C\}.
\]
For a cell $(c,h)$ let
$L_X=(\sum_i t_i^2)^{1/2}$. Every compatible state obeys
\begin{equation}
 \|y-X(c)x\|\le\sqrt C+A_1hL_X=:H_c.
 \label{rebuild:eq:cell-residual-transfer}
\end{equation}
This follows from the per-row rotation bound and the Frobenius bound on
$X(\nu)-X(c)$.

Let $G_c=X(c)^\top X(c)$, $b_c=X(c)^\top y$, and let
$F_c^{\rm lo}$ be a validated lower bound on the unrestricted profiled
loss $\inf_x\|y-X(c)x\|^2$. If $F_c^{\rm lo}>H_c^2$, the entire cell is
excluded. Otherwise, suppose $\lambda_c>0$ is a validated lower bound on
$\lambda_{\min}(G_c)$. The exact least-squares state is
$\widehat x_c=G_c^{-1}b_c$, and completing the square gives
\[
 \|x-\widehat x_c\|\le
 \sqrt{(H_c^2-F_c^{\rm lo})/\lambda_c}.
\]
If a numerical center $z_c$ has certified normal-equation residual
$\|G_c z_c-b_c\|\le\rho_c$, a valid ball radius is
\begin{equation}
 r_c=\sqrt{(H_c^2-F_c^{\rm lo})/\lambda_c}+\rho_c/\lambda_c.
 \label{rebuild:eq:complete-state-ball}
\end{equation}
The numerator and all radii are rounded outward. If positive definiteness
is unresolved, the original ball centered at zero with radius $A_1$
remains valid. Thus weak excitation, zero amplitude, and resonant
frequencies affect tightness without invalidating coverage.

An alternative uses the quadratic actually stored in floating-point
arithmetic, avoiding an assumption that its numerical minimizer lies in
the amplitude disk. Let
$q_c(x)=\widetilde C-2\widetilde b^\top x+x^\top\widetilde Gx$ and suppose
$|\|y-X(c)x\|^2-q_c(x)|\le a_c$ for every physical state
$\|x\|\le A_1$. Certify $g_c\le\lambda_{\min}(\widetilde G)$ with
$g_c>0$, $f_c\le q_c(z_c)$, and
$\rho_c\ge\|\widetilde Gz_c-\widetilde b\|$. Then the directly computed
ball
\begin{equation}
 r_c=\frac{\rho_c}{g_c}
   +\sqrt{\frac{\max\{0,H_c^2+a_c-f_c+\rho_c^2/g_c\}}{g_c}}
 \label{rebuild:eq:stored-quadratic-ball}
\end{equation}
contains every compatible physical state. Indeed, for $u=x-z_c$,
$q_c(x)\ge f_c-2\rho_c\|u\|+g_c\|u\|^2$, while compatibility gives
$q_c(x)\le H_c^2+a_c$. Solving this scalar quadratic yields the display.
A negative radicand implies that the cell has no compatible state, so replacing it by zero preserves containment.
This argument does not identify the stored quadratic's minimum with the
exact unrestricted profile. Its lower anchor $f_c-\rho_c^2/g_c$ must not
be clipped upward to zero when computing a state radius.
The same completion in the $\widetilde G$ norm gives the sharper ellipsoid
\[
 (x-z_c)^\top\widetilde G(x-z_c)\le
 \left\{\sqrt{\max\{0,H_c^2+a_c-f_c+\rho_c^2/g_c\}}
                 +\rho_c/\sqrt{g_c}\right\}^2=:Q_c^{\rm rad}.
\]
Its support in direction $v$ is at most
$v^\top z_c+\sqrt{Q_c^{\rm rad}\,v^\top\widetilde G^{-1}v}$.
This follows by completing the square with
$\widetilde G^{-1}(\widetilde Gz_c-\widetilde b)$ and using
$(\widetilde Gz_c-\widetilde b)^\top\widetilde G^{-1}
(\widetilde Gz_c-\widetilde b)\le\rho_c^2/g_c$.

The frequency cells initially cover the whole circle. A cell can be
removed only after a validated exclusion; every remaining cell is retained
or subdivided into a complete child cover. The resulting finite collection
and its state balls therefore meet Theorem~\ref{rebuild:thm:trajectory-certificate}.
If computation stops before a desired width is attained, keeping unresolved
cells with the full state disk still gives a valid, possibly loose bound.
The trajectory routine uses the deterministic bound when the cutoff is
negative, the cover is empty, or its requested refinement is unfinished.
It records this fallback separately from a completed polygon certificate.

\paragraph{Two oscillators.}
For plane $j$, let $\mathcal C_{j,T}$ be a separately covered state-frequency
confidence set. Allocate $\delta_1+\delta_2=\delta$ so both cover the truth
with probability at least $1-\delta$. For an action difference, let
$v_{t,A,j}$ collect that plane's coordinate coefficients. Summing the two
plane support bounds before maximizing over the same feasible pair $A$
gives a valid per-round regret bound. Summing in time gives a trajectory
certificate; keeping one cell or polygon vertex per plane across all
times gives a potentially tighter finite product bound. This extension
uses the actual feasible action family, not an independent best-arm
choice for each plane.

\paragraph{Coverage across repeated runs.}
For $N$ separately reported trajectories, using failure probability
$\delta$ for each gives per-trajectory coverage. It does not establish
simultaneous $1-\delta$ coverage of all $N$ trajectories. To obtain that
claim, allocate probabilities $\delta_1,\ldots,\delta_N$ with
$\sum_j\delta_j\le\delta$, and use the corresponding cutoff for each
trajectory. The allocation and the distinction between measured regret
and its validated upper bound are reported with the experiment protocol.

\subsection{A local nonlinear prediction comparison}
\label{sec:local-nonlinear-comparison}
The completed confidence cover also gives a deterministic check of local
observability.  This check concerns exact oscillator predictions, including
their nonlinear dependence on frequency.

Represent the state at the midpoint $t_0=n/2$ of the observation window,
and use the scaled parameter $\vartheta=(z_1,z_2,n\omega)$ in these
midpoint coordinates, with $n\ge1$. Frequencies use a continuous local
lift of the circle on the patch.  This removes the correlation introduced by
extrapolating the phase back to time zero.
For a patch with center $(z_c,\omega_c)$, suppose its states satisfy
$\|z-z_c\|\le r_z$ and its frequencies satisfy
$n|\omega-\omega_c|\le r_\omega$.
All physical states lie in the amplitude disk of radius $A$.
For an observation of coordinate $b$ at time $t$, put $q=(t-t_0)/n$ and
$d_t(\omega)=R(-(t-t_0)\omega)e_b$.
The Jacobian of its mean with respect to $\vartheta$ has the row
\[
 J_t(\vartheta)=\bigl(d_t(\omega)^\top,\;q\,\widetilde d_t(\omega)^\top z\bigr),
\]
where $\widetilde d_t(\omega)=\left.\partial_u[R(-u)e_b]\right|_{u=(t-t_0)\omega}$
is the angle derivative of the sensing vector. Let $J$ stack these rows and $\vartheta_c=(z_c,n\omega_c)$ be the patch center. Throughout the patch,
\[
 \|J(\vartheta)-J(\vartheta_c)\|_{\rm op}\le D,\qquad
 D^2=(r_z^2+r_\omega^2)\sum q^2
 +2Ar_zr_\omega\sum |q|^3+A^2r_\omega^2\sum q^4.
\]
The sums run over scalar rows, including repeated observation times.
This follows row by row from
$\|d_t(\omega)-d_t(\omega_c)\|\le |q| r_\omega$ and the derivative-column
bound $|q|(r_z+A|q| r_\omega)$.
The straight segment joining two states in the amplitude disk remains
in the disk; the frequency interval and state ball are also convex.

Let $s_-$ lower-bound the smallest singular value of the center Jacobian
for the selected observations through round $n$, and let $s_+$ upper-bound the
largest singular value for the two-coordinate predictions at rounds
$n+1,\ldots,2n$.  Let $D_-$ and $D_+$ be the corresponding bounds above.
If $s_->D_-$, integrating the Jacobian along the segment between any
two physical models in the patch gives
\[
 \sum_{t=n+1}^{2n}\|R((t-t_0)\omega)z-R((t-t_0)\omega')z'\|^2
 \le \left(\frac{s_++D_+}{s_--D_-}\right)^2
 \sum_{r:\,t_r\le n}\bigl(d_r(\omega)^\top z-d_r(\omega')^\top z'\bigr)^2.
\]
The scalar-row index $r$ runs over every selected physical observation,
with its round denoted by $t_r$.
For numerical verification, a complete confidence cover is enclosed in
local patches, its state radii are rounded outward, and the two $3\times3$
Jacobian Gram matrices are enclosed entrywise.  An arbitrary numerical
eigenvector matrix supplies a congruence; interval Gershgorin bounds on
that congruence and on its Gram matrix certify the singular values
without assuming exact orthogonality.  Different patches receive
separate certificates.

\subsection{Certifying that another sensing block costs more}
\label{app:reconstructed-cost-gate}

Let $S_t$ be the frozen greedy schedule proposed at checkpoint $n$.
Let $F_t$ be the public full-vector sensing schedule on the same future
rounds: both coordinates of plane 1 for two rounds, then both coordinates
of plane 2 for two rounds, repeated. Let $m_{\rm plan}=\min\{n,T-n\}$ be the
planned block length. Compare the schedules over $m=\min\{256,m_{\rm plan}\}$ rounds. Their reward difference under candidate
$\theta$ is
\[
 G_n(\theta)=\sum_{t=n+1}^{n+m}
 \left\{\sum_{i\in S_t}\mu_i^\theta(t)
              -\sum_{i\in F_t}\mu_i^\theta(t)\right\}.
\]
A positive validated lower bound on
$\inf_{\theta\in\mathcal C_{1,n}\times\mathcal C_{2,n}}G_n(\theta)$
certifies that the committed schedule earns more expected reward than
another sensing block. This comparison uses the complete confidence
class and is computed before observing any reward in the compared block.
It remains valid for off-grid true frequencies.

\begin{proposition}[Prospective sensing-cost and information gates]
\label{prop:reconstructed-cost-gate}
On a simultaneous confidence event, suppose the controller validates
$\underline G_n>0$ as a lower bound on the preceding infimum. Then the
proposed schedule has mean reward at least $\underline G_n$ greater
than the sensing schedule over these same $m$ rounds.
For plane $j$, let $d_{j,t,b}(\nu)=R(-t\nu)e_b$ denote its sensing column.
For a supplied threshold $\gamma>0$, if the controller also validates, for every retained frequency cell,
\[
 \lambda_{\min}\left(
 \sum_{n<t\le n+m_{\rm plan}:\,(j,b)\in S_t}
 d_{j,t,b}(\nu)d_{j,t,b}(\nu)^\top\right)\ge \gamma m_{\rm plan},
 \qquad j=1,2,
\]
then its planned observations give both true states information in every
direction, with least eigenvalue at least $\gamma m_{\rm plan}$.
\end{proposition}
\begin{proof}
Substitute the true pair of models into the two infima. Confidence
coverage places its frequencies and states inside the retained cover.
Both inequalities are deterministic conditional on the past and the
chosen schedules.
\end{proof}

At frequency-cell center $\nu_c$, let $Q_c,b_c,C_c$ be stored quadratic
coefficients and let $\widehat x_c$ be a numerical least-squares center.
Define $q_c(x)=C_c-2b_c^\top x+x^\top Q_cx$.
Let $C\ge0$ be the selected-loss confidence cutoff. For cell half-width $h$, put
$H_c=\sqrt C+A_{\max}h\sqrt{\sum t^2}$, summing over that plane's selected scalar observations, with repeated times counted.
A validated coefficient envelope $E_c$ ensures
$q_c(x)\le H_c^2+E_c$ for every compatible amplitude-bounded state.
Let $g_c>0$ be a validated lower bound on $\lambda_{\min}(Q_c)$,
let $q_c^-\le q_c(\widehat x_c)$ be evaluated with outward arithmetic,
and let $\rho_c\ge\|Q_c\widehat x_c-b_c\|$.
Completing the quadratic gives the valid ellipsoid
\[
 (x-\widehat x_c)^\top Q_c(x-\widehat x_c)\le B_c,\qquad
 \sqrt{B_c}=\frac{\rho_c}{\sqrt{g_c}}+
 \sqrt{\max\{0,H_c^2+E_c-q_c^-+\rho_c^2/g_c\}}.
\]
Indeed, writing $p^2=(x-\widehat x_c)^\top Q_c(x-\widehat x_c)$
yields $p^2-2\rho_cp/\sqrt{g_c}\le H_c^2+E_c-q_c^-$.
The center can lie outside the amplitude disk; its quadratic value is
evaluated directly, so the bound remains valid there.
A linear functional $v^\top x$ is consequently bounded above by
$v^\top\widehat x_c+\sqrt{B_c\,v^\top Q_c^{-1}v}$.
An unresolved positive Gram bound uses the physical disk support
$A_{\max}\|v\|$.
For the cost comparison, collect all rows of a plane into one vector
$v(\nu)=\sum_t\sum_b a_{t,b}R(-t\nu)e_b$, where
$a_{t,b}=1$ for a sensing-only arm, $-1$ for a committed-only arm,
and zero otherwise. The interval variation is bounded by
$A_{\max}h\sum_t t\sum_b|a_{t,b}|$.
Maximizing these bounds over every retained cell and adding both planes
gives an outward upper bound on $-G_n$, hence a lower bound on $G_n$.
Aggregating before taking the support function preserves cancellations
between rounds and makes the prospective cost comparison substantially
sharper than summing separate forecast widths.

For the information gate, center times at the block midpoint $t_c$.
Let $Q^{\rm plan}_{j,c}$ be the Gram matrix of the planned sensing columns at cell center $\nu_c$, rotated to this midpoint origin. Orthogonal conjugation preserves eigenvalues. Since the derivative of
the rank-one projector $R(-s\nu)e_be_b^\top R(s\nu)$ has operator norm
$|s|$, a frequency cell has certified Gram floor
\[
 \lambda_{\min}(Q^{\rm plan}_{j,c})
       -h\sum_{t,b:\,(j,b)\in S_t}|t-t_c|.
\]
The minimum over all retained cells is the reported plane-wise floor.
All retained intervals cover the original frequency circle's compatible
models; numerical proposal searches do not remove intervals.

\section{Numerical implementation}
\label{app:structural-current-numerics}
\subsection{State supports for a finite confidence cover}
\label{app:ridge-support}
A frequency interval can be useful before its width reaches a prescribed
mesh tolerance. The numerical controller retains every interval that has
not been excluded by a certified loss lower bound, including every interval
remaining at the evaluation cap. These retained intervals form an outer
cover of the confidence set.

Fix one retained interval with center $\nu$ and half-width $h$.
Let $D_\nu$ denote the selected-observation design, let $\ell\ge0$ be the
confidence loss cutoff, and assume $\|z\|\leq A$.
For every feasible frequency in the interval,
\[
 \|D_\nu z-y\|\leq \sqrt\ell+Ah\sqrt{\textstyle\sum_i t_i^2}.
\]
Write $H_\nu=\sqrt\ell+Ah\sqrt{\sum_i t_i^2}$ for this residual radius.
The stored coefficients $Q,b,C$ represent the exact-real quadratic
$q(z)=C+z^\top Qz-2b^\top z$.
The profile oracle supplies an outward coefficient envelope $E$ such that
$|q(z)-\|D_\nu z-y\|^2|\leq E$ throughout the physical amplitude disk.
Thus every feasible state satisfies
\begin{equation}
 z^\top Qz-2b^\top z\leq H_\nu^2+E-C.
 \label{eq:ridge-loss-sublevel}
\end{equation}

Write $\operatorname{fl}$ for floating-point evaluation. Choose any $\lambda\geq0$ for which the computed symmetric matrix
$M=\operatorname{fl}(Q+\lambda I)$ has a verified eigenvalue lower bound
$g>0$. Let $F$ enclose $|z^\top(M-Q-\lambda I)z|$ on the amplitude disk.
For any proposed center $x$, let $\rho\geq\|Mx-b\|$, and let
$q_x^-$ be a lower bound on $x^\top Mx-2b^\top x$.
Define
\[
 B=H_\nu^2+E-C+\lambda A^2+F-q_x^-+2\rho(A+\|x\|).
\]
Expanding the quadratic around $x$ and using
$\|z-x\|\leq A+\|x\|$ gives
\[
 (z-x)^\top M(z-x)\leq B.
\]
Consequently, for any contrast $v$, an upper state support is
\begin{equation}
 v^\top x+\sqrt{\max\{B,0\}\,v^\top M^{-1}v}.
 \label{eq:ridge-state-support}
\end{equation}
A numerical inverse proposal $u$ is sufficient: if
$r\geq\|Mu-v\|$, then
$v^\top M^{-1}v\leq v^\top u+\|v\|r/g$.
All displayed numerical quantities receive outward arithmetic envelopes.
The implementation also uses the independent disk support $A\|v\|$ and
takes the minimum over these valid supports and a fixed list of nonnegative
ridge values. No invertibility assumption on the original selected design
is needed.

For a future contrast at time $t$, replacing the unknown frequency by
$\nu$ adds at most $A\|v\|\min\{2,|t|h\}$, together with the bounded
trigonometric evaluation error. Maximizing over every retained interval
and taking the positive part gives a per-round regret bound. An outward
sum supplies the held-block bound. These bounds are valid whether or not
the requested mesh tolerance was achieved.

\paragraph{Adaptive block ends.}
After computing the bounds for a proposed block, the controller first tests
its full length. If that test fails, it checks every shorter integer prefix
using that prefix's own budget
$c\sigma\sqrt{\log(8T/\delta)}(\sqrt m-\sqrt n)$ and selects the
longest certified prefix. If no prefix passes, it executes four active
fallback rounds, or the remaining horizon when shorter, and checks again.
All endpoints depend only on selected observations already available.
The accepted intervals are disjoint, so their square-root budget increments
sum to at most $c\sigma\sqrt{T\log(8T/\delta)}$.
The fixed-prior confidence event is simultaneous over these adaptive checks.

\subsection{State radii implied by the selected loss}
\label{app:local-enclosure}
A frequency interval can contain only low-amplitude models even when the supplied amplitude bound is large. The selected loss gives a uniform way to recover this information before computing future regret. This also sharpens the loss bound used to exclude intervals.

Write the selected loss as $L(\nu,z)=C-2b(\nu)^\top z+z^\top Q(\nu)z$. For an interval $I=[c-h,c+h]$, let $g_I$ lower-bound the smallest eigenvalue of $Q(\nu)$ throughout $I$, let $\beta_I$ upper-bound $\|b(\nu)\|$ throughout $I$, and let $C_-\le C$. Every state of radius $r=\|z\|$ then satisfies
\begin{equation}\label{eq:local-radial-loss}
 L(\nu,z)\ge C_--2\beta_I r+g_I r^2,\qquad \nu\in I.
\end{equation}
These coefficient bounds use all selected observations. If their physical times and rewards are $t_u,y_u$, define $S_2=\sum_u t_u^2$ and $V_2=\sum_u t_u^2|y_u|$. Rotation derivatives give
\[
 \begin{aligned}
 \|Q(c+v)-Q(c)-vQ'(c)\|&\le h^2S_2,\\
 \|b(c+v)-b(c)-vb'(c)\|&\le\tfrac12h^2V_2,\qquad |v|\le h.
 \end{aligned}
\]
The smallest eigenvalue is concave along a matrix segment, and the vector norm is convex along a vector segment. Thus valid exact-arithmetic choices are
\[
\begin{split}
 g_I&=\min_{v\in\{-h,h\}}\lambda_{\min}(Q(c)+vQ'(c))-h^2S_2,\\
 \beta_I&=\max_{v\in\{-h,h\}}\|b(c)+vb'(c)\|+\tfrac12h^2V_2.
\end{split}
\]
The implementation subtracts its outward coefficient-error envelopes from $g_I$ and adds them to $\beta_I$. It also rounds $C_-$ downward. These envelopes are independent of the unknown state's amplitude.

Let $\ell$ be the selected-loss cutoff defining compatible models in this query. If $g_I>0$, compatibility and \eqref{eq:local-radial-loss} imply the upper radius
\begin{equation}\label{eq:local-radius}
 \|z\|\le \rho_I
 :=\min\left\{A_{\max},\,
 \frac{\beta_I+\sqrt{\beta_I^2-g_I(C_--\ell)}}{g_I}\right\}.
\end{equation}
For $g_I\le0$, set $\rho_I=A_{\max}$. A negative discriminant implies that no compatible state exists. A nonfinite or unresolved arithmetic comparison also retains $\rho_I=A_{\max}$. The radius is used only for cutoffs no larger than the one under which it was established.

The radial polynomial also supplies a uniform loss lower bound. For $g_I>0$, its unrestricted minimum is $C_--\beta_I^2/g_I$. For $g_I\le0$, it decreases on $[0,A_{\max}]$, so its minimum there is $C_--2\beta_I A_{\max}+g_I A_{\max}^2$. Taking the maximum with the existing loss lower bound preserves validity. Every surviving model obeys \eqref{eq:local-radius}, allowing $\rho_I$ to replace $A_{\max}$ in the state support and frequency-drift bounds of the preceding subsection. The minimum of the original and sharpened support bounds is again valid. At the refinement cap, every nonexcluded interval is still retained.

\subsection{Arithmetic and unresolved comparisons}
\label{app:coefficient-contract}\label{app:extended-coefficients}
The mathematical inequalities above concern real-valued quadratics and their support functions. The implementation encloses their coefficient, trigonometric, residual, and summation errors under IEEE binary64 round-to-nearest arithmetic with gradual underflow. The selected-loss oracle evaluates bounded polynomial sine and cosine approximations after range reduction, rather than assuming an unspecified library trigonometric error.

For $N$ selected scalar observations, the accumulation envelope uses
\[
 \gamma_{8N}=\frac{8Nu}{1-8Nu},\qquad u=2^{-52}.
\]
The canonical entry points require $1\le N<2^{49}$ before allocation, ensuring a positive denominator. Each planned support sum has its own corresponding summation check. Unsupported arithmetic or a failed confidence computation produces a recorded decline. The controller then executes its stated online-learning continuation.

Complete coverage and requested numerical resolution are recorded separately. At a refinement cap, all nonexcluded intervals remain in the cover. Their conservative supports continue to bound regret, potentially leading to a shorter held block or a learning decision. A query can therefore stop after a finite computation budget without treating unsearched parameter regions as absent.

A verified feasible countermodel can also prove that every candidate prefix exceeds its budget. This recorded witness permits the implementation to skip support evaluations for that refusal. Identical loss-only interval refinement preserves the retained domain and subsequent decisions. The witness, feasibility check, and all-prefix regret margins are saved for replay.

The empirical audits compare physical losses, support evaluations, and posterior calculations with independent implementations, including rank-deficient and resonant cases. These checks test the numerical realization of the stated arithmetic contract. The source preserves coefficient enclosures and decision logs so the resulting acceptances can be replayed.

\section{Experiments and reproducibility}
\label{app:exp}
\subsection{Selected feedback and paired potential rewards}
\label{app:reconstruction-protocols}
Every physical arm has a potential Gaussian noise value for each round, generated before a run with NumPy's PCG64 generator and the declared seed. On a paired instance all policies share this array. Both selected arms are fixed before either current reward is exposed. Numerical fitting receives only the selected physical times, coordinate labels, and rewards. True states and unselected rewards enter the separate evaluation and audit code.

Each run stores its policy, environment, seed, source hashes, potential noise, selected actions and rewards, numerical fits, confidence checks, regret increments, and completion status. A failed numerical comparison selects the recorded learning continuation; a stopped run retains its partial status. The figures and tables read completed records for their declared cohorts. Earlier versions and unsuccessful attempts have separate directories and remain traceable.

\subsection{Primary catalog and numerical fits}
The twelve primary cells use frequencies
\[
\begin{split}
&(0.508634,0.808848,1.058562,2.079430,2.379644,2.629359,\\
&\hspace{8pt}3.650227,3.950441,4.200155,5.221023,5.521237,5.770951),
\end{split}
\]
printed to six decimals here; the frozen JSON protocol contains the complete binary64 values. Amplitudes range over $1.021,1.247,1.479$ and noise scales over $0.05,0.10,0.20$. The phases and allocation of these values to cells are fixed in that protocol. Seed $73{,}000{,}000+1000c+r$ specifies one-based cell $c$ and replicate $r\in\{0,\ldots,15\}$. All primary trajectories have $T=1024$.

Every proposal fit searches $\omega\in[0,2\pi)$ and $\|z\|\le1.5$. The public initial model is $(\omega,z)=(0.4949341263408955,(1,0))$. At a fixed frequency, a two-dimensional constrained least-squares solve profiles out the state. Let $t_{\max}$ be the last selected observation time. A fast Fourier transform (FFT) scan uses the next power of two at least $\max\{64,8(t_{\max}+1)\}$. The eight best local minima from each of a lower-score and an upper-score scan seed bounded continuous refinements. Interval endpoints and the preceding fitted frequency are also evaluated. Frequency refinement uses tolerance $2\cdot10^{-12}$ and at most seventy iterations. Objective ties within $2\cdot10^{-11}\max\{1,\|y\|^2\}$ prefer the frequency nearest the incumbent, then the lower frequency. These numerical proposals are separate from the complete confidence bounds.

The full-disk state solver handles rank-deficient endpoint Gramians by the analytic zero-denominator limit of its secular equation. Independent rank-one checks compare its state and loss with direct constrained optimization. The repaired engine is versioned separately from the original restricted engine.

Dense NLS refits at every nonempty selected history. The sparse control refits through the 34-round prefix, then at its declared checkpoint schedule. An initial mixed pair uses physical arms 0 and 2. All subsequent one-plane greedy choices take a replica pair, with the horizontal coordinate winning exact action ties.

\subsection{Checked blocks and the active learning continuation}
For one plane, the default nominal checkpoint multiplier is two, the prefix is 34, and $c=16$ in \eqref{eq:gate-main}. The confidence calculation receives $\delta/2$ with $\delta=0.05$. A whole-interval search uses at most 16,384 refinement evaluations. Intervals that remain unresolved are retained in the confidence cover. Its regret upper bound therefore continues to cover every compatible parameter, even when broad intervals remain.

If the full nominal schedule fails the budget, conservative per-round upper bounds from the same retained cover are accumulated for every integer prefix. The longest prefix satisfying its own budget is executed. If no prefix is certified, the learner takes four active SquareCB rounds, or the remaining horizon if shorter, and refits. The next nominal endpoint is computed from this actual endpoint. The online regression banks train only on their own active rounds, using physical time as the prediction context. They are separate from the all-history NLS proposal and confidence calculation.

The canonical entry points record arithmetic failures as declined checks, retain the preceding valid domain, and follow the learning continuation. Before allocating history arrays they reject a selected-row count at least $2^{49}$, outside their declared binary64 accumulation contract. Numerical checks do not silently replace an unresolved comparison by an acceptance. The source and audit records identify the confidence and arithmetic implementation used by every run.

\subsection{Oscillator posterior and disjoint development}
The independent Thompson sampler assigns equal prior mass to frequencies $2\pi k/F$, $k=0,\ldots,F-1$, with $F=262144$. The initial state has prior $N(0,s^2I_2)$. At each frequency the selected-data Gaussian likelihood integrates the two-dimensional state analytically, including the frequency-dependent determinant normalization. Sampling first draws a frequency from these masses, then draws its conditional Gaussian state. The resulting model determines the greedy action.

A pure pair is averaged and enters the likelihood with noise variance $\sigma^2/2$. A first-round mixed pair supplies separate horizontal and vertical observations with variance $\sigma^2$ each. The model draw for that round is still consumed before the mixed action overrides it. This preserves the declared posterior random stream across the two schedules. The state prior is unbounded, whereas the confidence controller uses the supplied amplitude disk.

Development seeds $83{,}000{,}000+1000c+r$, $r=0,1,2$, evaluate prior scales $s\in\{0.5,1,1.5\}$ on all twelve cells. The lowest unrounded equal-cell mean selects $s=0.5$. Its development mean is 4.06035, compared with 4.23846 and 4.11097 for scales one and 1.5. A prespecified resolution comparison uses one development seed per cell at $F=65536,131072,262144$, yielding means 5.01472, 4.99559, and 4.96575. The largest grid is fixed for evaluation.

Posterior randomness uses PCG64 with a seed sequence containing the environment seed and two fixed stream identifiers. Trigonometric recurrences reset every 64 rounds, binary64 likelihoods are normalized at each round, and no frequency is pruned. The independent audit checks the marginal likelihood and conditional state moments against direct selected-history matrix calculations.

\subsection{Fresh phase coverage and checkpoint confirmation}
The confirmation study keeps the twelve primary frequency, amplitude, and noise combinations. For zero-based cell index $i$, it sets the first physical phase to $2\pi(i+0.37)/12$ and the initial phase to that angle minus the cell's frequency modulo $2\pi$. This gives three first-observation phases in each quadrant. It is a fixed twelve-cell design, rather than a full frequency-by-phase factorial.

Eight new evaluation seeds $91{,}300{,}000+1000c+r$, $r=0,\ldots,7$, give 96 instances. The frozen comparison includes nominal growth factors 1.25 and two for checked commitment, matched sparse NLS at 1.25, dense NLS, and the independently selected posterior. Every method takes one mixed initial observation. Both checkpoint settings are reported. The source package preserves the protocol checksum before the first outcome and the execution versions applied to it.

\subsection{Trajectory bounds and paired uncertainty}
Terminal certificates use the first four seeds from each primary cell, separately for zero-mixed and one-mixed controllers: 48 trajectories per schedule and 96 in total. Each receives its own failure probability 0.05. These are individual trajectory guarantees. The confidence prior has uniform frequency and Gaussian state scale 1.5; its confidence set is intersected with the disk of radius 1.5. The first two seeds per fresh confirmation cell supply 24 separately declared certificates for the shorter nominal schedule. Displayed decimal bounds are rounded upward.

Reported paired uncertainty resamples seed replicates independently within each fixed cell, preserving matched methods and assigning equal weight to cells. Ten thousand bootstrap draws and a fixed analysis seed give the intervals. The two posterior comparisons in each cohort use 97.5\% intervals for a nominal Bonferroni family level of 95\%; other intervals are labeled pointwise 95\%. This measures simulation noise on the declared environments. Terminal certificates instead make a probability statement about a trajectory under its noise model; the two uses of probability have different meanings.

\subsection{Finite uniform bounds and computational accounting}
At $T=1024$, $A_{\max}=1.5$, $n_0=34$, $c=16$, and $\delta=0.05$, the explicit bound of Theorem~\ref{thm:reconstructed-main} is at most 2427, 2682, and 3192 for noise $0.05$, $0.10$, and $0.20$. For the primary two-plane catalog, $T=8192$, $A_{\max}=1.5$, $\sigma=0.25$, $n_0=256$, and per-trajectory failure probability $0.05/16$. Its budget coefficient is $c=39.414$ (rounded here). The finite uniform bound is at most 29,610, compared with the deterministic maximum 34,756. These uniform coefficients are conservative; the terminal bounds use the selected observations to quantify individual trajectories.

The complete records supply exact nonlinear-refit counts: 1023 for dense NLS, 38 for the primary doubling schedule, and 49 for the fresh growth-factor-1.25 schedule. On these catalogs, the checked policy and its matched sparse control use the same refit cadence. These counts are independent of hardware and include every trajectory in the reported catalogs.

\newcommand{\RevTimingRows}{Dense NLS & 7.021 & 7.057 & 7.015 & 1023 / 1023 \\
Checked commitment & 0.648 & 0.559 & 1.515 & 38 / 49 \\
Matched sparse NLS & 0.230 & 0.178 & 0.227 & 38 / 49 \\
Oscillator posterior & 5.475 & 5.571 & 5.441 & -- \\
Checked, fresh doubling & -- & -- & 0.588 & -- / 38 \\}
\newcommand{\RevTimingComparison}{The median paired dense-to-checked runtime ratio is 10.81 on the primary zero-mixed catalog and 4.69 on fresh mixed-start instances. The corresponding sparse-to-checked ratios are 0.35 and 0.16. These ratios include confidence computation; the sparse comparison measures its overhead at matched fitting cadence. Ratios above one favor checked commitment.}

\label{app:runtime}
The frozen isolated timing study uses one prespecified seed in each of the
twelve cells: 96 primary executions across eight method/sensing combinations
and 60 fresh executions across five methods. Methods run sequentially in a
fixed randomized order, with one numerical-library thread and no concurrent
experiment, build, rendering, or compression job. Native compilation and one
64-round warmup per named method occur outside the measured clocks.

The core clock includes environment generation and action computation.
The common end-to-end clock adds compression of the same five trace arrays,
their SHA256 digest, and the same metadata schema for every method. It
excludes research archive serialization. After measurement, all five arrays
must match the method's production trajectory exactly. Every measured
execution passes this check. Confidence refinement and block checks are
included in the checked controller's clock. These are measurements on one
execution environment; they do not imply hardware-independent speedups.

\begin{table}[!htbp]
\centering\small
\setlength{\tabcolsep}{4pt}
\begin{tabular}{lrrrr}
\toprule
& \multicolumn{2}{c}{Primary seconds} & Fresh seconds & Refit counts \\
Method & Zero mixed & One mixed & One mixed & Primary / fresh \\
\midrule
\RevTimingRows
\bottomrule
\end{tabular}
\caption{Median common end-to-end runtime in seconds over the twelve
prespecified instances in each column. The first checked row uses doubling
on the primary catalog and growth factor 1.25 on the fresh catalog; the last
row gives the separately prescribed fresh doubling schedule. Every
trajectory has horizon 1024. Refit counts do not include confidence work.}
\label{tab:runtime-fresh}
\end{table}
Table~\ref{tab:runtime-fresh} summarizes the timings. \RevTimingComparison{}
Matched sparse NLS is faster than checked commitment in all three
matched-cadence timing comparisons.

\subsection{Two-plane protocol}
The two-plane experiment uses frequency pairs $(0.73,0.76)$ and $(0.73,1.17)$, amplitudes $(1,1)$, phases $(0,0.7)$, noise $0.25$, horizon 8192, and seeds $27001,\ldots,27008$. The controller and matched NLS share potential rewards for each instance. Both coordinates of plane one are read in rounds 1--2 and both coordinates of plane two in rounds 3--4. Useful selected-feedback learning continues through round 256, followed by the declared checkpoint rule.

Two-plane endpoints are fixed at $\min\{2n,T\}$. A declined check runs the active learner to that endpoint. The cost gate compares the proposed greedy schedule with the public dedicated sensing schedule on the same next 256 rounds, shortened at the horizon. The information gate bounds both planned Gramians over all retained frequencies. The regret budget is the telescoping gate in \eqref{eq:gate-main}. Failure probability $0.05$ is divided among the sixteen trajectories. The exact budget conversion and online-learning constants are specified in Appendix~\ref{app:active-rate}. Figure~\ref{fig:sensing-cost-supplement} shows the cost bounds; the main text reports the executed learning continuation.

\begin{figure}[!ht]
\centering
\includegraphics[width=\linewidth]{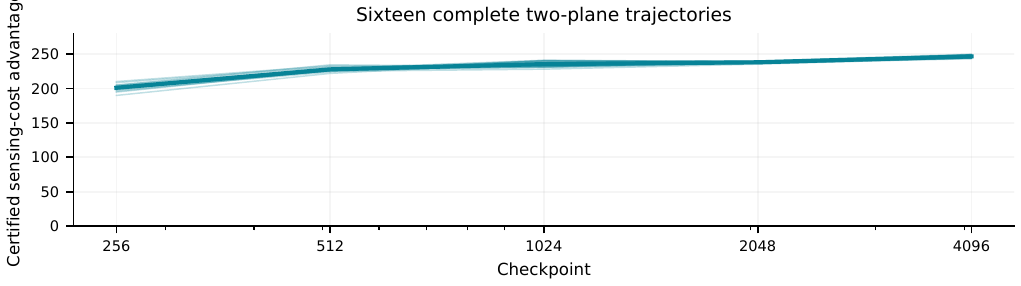}
\caption{Prospective two-plane reward advantage over the specified additional sensing schedule, certified before each block. Thin lines show the sixteen complete primary trajectories; the thick line is their mean. Every lower bound is positive.}
\label{fig:sensing-cost-supplement}
\end{figure}

\subsection{Exact and near-resonant stress evaluation}
\label{app:stress-fresh}
The fixed design crosses frequencies $0$, $0.01$, $\pi/2$, and
$\pi/2+0.01$ with amplitudes $0.15$ and $1.2$ at $T=1024$.
Noise is $0.05$ at exact resonance and $0.2$ near resonance.
The first physical phase is $0$ at the lower amplitude and $\pi/4$ at the
higher amplitude. Seed $97{,}500{,}000+1000c+r$ specifies cell
$c\in\{1,\ldots,8\}$ in Table~\ref{tab:stress-fresh}.
Replicates $r=0,\ldots,3$ form the previous cohort and $r=4,5$ the original
holdout cohort; these seeds were fixed before the final enclosure implementation.

We add matched unchecked sparse NLS to this eight-cell comparison.
The methods share potential noise, initialization, numerical fitting,
and greedy tie rules. The sparse control fits after histories $1,\ldots,33$
and then at $34,68,136,272,544$, giving 38 fits. Checked commitment
uses the same nominal doubling rule and chooses its executed endpoints
through the budget checks. Table~\ref{tab:stress-fresh} reports all three
methods and both original cohorts.

The paired checked-minus-sparse mean regret is
$\ReviewStressPreviousPairedMean$ on the previous cohort and
$\ReviewStressHoldoutPairedMean$ on the original holdout cohort.
The respective counts of checked wins, ties, and losses are
$\ReviewStressPreviousWinTieLoss$ and $\ReviewStressHoldoutWinTieLoss$,
using absolute tie tolerance $10^{-10}$.
Previous checked runs total \ReviewStressPreviousLearningRounds{} learning
rounds and \ReviewStressPreviousShortenedBlocks{} shortened blocks;
the original holdout totals are \ReviewStressHoldoutLearningRounds{}
and \ReviewStressHoldoutShortenedBlocks{}.

\setlength{\textfloatsep}{10pt plus 2pt minus 2pt}
\setlength{\intextsep}{10pt plus 2pt minus 2pt}
\begin{table}[!htbp]
\centering\small
\setlength{\tabcolsep}{2.6pt}
\begin{tabular}{llrrrrrrrrr}
\toprule
& & & \multicolumn{3}{c}{Previous (4/cell)}
& \multicolumn{3}{c}{Original holdout (2/cell)}
& \multicolumn{2}{c}{Holdout totals} \\
Cell & $\omega$ & $A$ & C & S & D & C & S & D & F & B \\
\midrule
\ReviewStressRows
\bottomrule
\end{tabular}
\caption{Stress regret and nonlinear refits. C: checked commitment;
S: matched unchecked sparse NLS; D: dense NLS. Cell rows give mean regret.
F and B count C's active-learning rounds and shortened blocks across the
two original holdout seeds. Final rows summarize all 32 previous or
16 original holdout instances; percentiles use linear interpolation.}
\label{tab:stress-fresh}
\end{table}

\paragraph{Full execution cost.}
Table~\ref{tab:stress-resources} compares all sixteen original holdout
stress instances and both early two-plane instances. The latter use
frequencies $(0.91,0.94)$, noise $0.25$, horizon $8192$, and a first check
at round 4. Runs use fresh processes, one numerical-library thread,
and a fixed randomized sequential order; 64-round warmups run separately.
Time includes environment generation, all decisions, and the same
five-array compressed payload with hashes and metadata.

Imports and identity audits are outside the clock. Peak resident set size (RSS) is
measured after payload writing and before the identity audit; it includes
imports and the full regression banks when allocated. The banks contain
59,479 frequencies for one plane and 1,344,975 per bank for two planes.

\begin{table}[!htbp]
\centering\small
\setlength{\tabcolsep}{3pt}
\begin{tabular}{llrrrrrr}
\toprule
& & \multicolumn{2}{c}{Checked}
& \multicolumn{2}{c}{Sparse} & \multicolumn{2}{c}{Dense} \\
Case & Replicate/seed & Seconds & MiB & Seconds & MiB & Seconds & MiB \\
\midrule
\ReviewResourceRows
\midrule
\multicolumn{8}{l}{Cohort summaries: mean seconds and maximum process MiB} \\
\ReviewResourceSummaryRows
\bottomrule
\end{tabular}
\caption{Isolated common end-to-end runtime and absolute process peak
memory. R01--R08 refer to Table~\ref{tab:stress-fresh}; replicates 4 and 5
are the original holdout seeds. The two-plane rows compare checked
commitment with matched NLS at the same
nominal checkpoints; dense two-plane execution is not part of this
comparison. \ReviewEarlyControlComparison{}
A MiB is $2^{20}$ bytes.}
\label{tab:stress-resources}
\end{table}


\begin{thebibliography}{22}

\bibitem[Abbas et~al.(2016)Abbas, Sun, and Foroosh]{abbas2016nonuniform}
Syed Alam Abbas, Qiyu Sun, and Hassan Foroosh.
\newblock Frequency estimation of sinusoids from nonuniform samples.
\newblock \emph{Signal Processing}, 129:67--81, 2016.
\newblock doi: 10.1016/j.sigpro.2016.05.024.

\bibitem[Azoury and Warmuth(2001)]{azoury2001relative}
Katy S. Azoury and Manfred K. Warmuth.
\newblock Relative loss bounds for on-line density estimation with the exponential family of distributions.
\newblock \emph{Machine Learning}, 43(3):211--246, 2001.
\newblock doi: 10.1023/A:1010896012157.

\bibitem[Bastani et~al.(2021)Bastani, Bayati, and Khosravi]{bastani2021greedy}
Hamsa Bastani, Mohsen Bayati, and Khashayar Khosravi.
\newblock Mostly exploration-free algorithms for contextual bandits.
\newblock \emph{Management Science}, 67(3):1329--1349, 2021.
\newblock doi: 10.1287/mnsc.2020.3605.

\bibitem[Chen et~al.(2025)Chen, Wang, and Wang]{chen2025seasonal}
Ningyuan Chen, Chun Wang, and Longlin Wang.
\newblock Learning and optimization with seasonal patterns.
\newblock \emph{Operations Research}, 73(2):894--909, 2025.
\newblock doi: 10.1287/opre.2023.0017.

\bibitem[Davis and Bucher(2019)]{davis2019frequency}
Solomon Davis and Izhak Bucher.
\newblock A new approach to single-tone frequency estimation via linear least squares curve fitting.
\newblock arXiv preprint arXiv:1904.07572, 2019.
\newblock URL \url{https://arxiv.org/abs/1904.07572}.

\bibitem[Emmenegger et~al.(2023)Emmenegger, Mutn\'{y}, and Krause]{emmenegger2023likelihood}
Nicolas Emmenegger, Mojm\'{i}r Mutn\'{y}, and Andreas Krause.
\newblock Likelihood ratio confidence sets for sequential decision making.
\newblock In \emph{Advances in Neural Information Processing Systems}, volume~36, 2023.
\newblock URL \url{https://arxiv.org/abs/2311.04402}.

\bibitem[Foster and Rakhlin(2020)]{foster2020beyond}
Dylan~J. Foster and Alexander Rakhlin.
\newblock Beyond UCB: Optimal and efficient contextual bandits with regression oracles.
\newblock In \emph{Proceedings of the 37th International Conference on Machine Learning}, volume~119 of \emph{Proceedings of Machine Learning Research}, pp.\ 3199--3210. PMLR, 2020.
\newblock URL \url{https://proceedings.mlr.press/v119/foster20a.html}.

\bibitem[Garivier and Moulines(2011)]{garivier2011ucb}
Aur\'{e}lien Garivier and Eric Moulines.
\newblock On upper-confidence bound policies for switching bandit problems.
\newblock In \emph{Proceedings of the 22nd International Conference on Algorithmic Learning Theory}, volume~6925 of \emph{Lecture Notes in Computer Science}, pp.\ 174--188. Springer, 2011.

\bibitem[Gornet and Sinopoli(2024)]{gornet2024restless}
Jonathan Gornet and Bruno Sinopoli.
\newblock Restless bandits with rewards generated by a linear Gaussian dynamical system.
\newblock In \emph{Proceedings of the 6th Annual Learning for Dynamics \& Control Conference}, volume~242 of \emph{Proceedings of Machine Learning Research}, pp.\ 1791--1802. PMLR, 2024.
\newblock URL \url{https://proceedings.mlr.press/v242/gornet24a.html}.

\bibitem[Gornet et~al.(2025)Gornet, Mo, and Sinopoli]{gornet2025kode}
Jonathan Gornet, Yilin Mo, and Bruno Sinopoli.
\newblock An exploration-free method for a linear stochastic bandit driven by a linear Gaussian dynamical system.
\newblock In \emph{Proceedings of the 64th IEEE Conference on Decision and Control}, pp.\ 5493--5500, 2025.
\newblock doi: 10.1109/CDC57313.2025.11312563.
\newblock URL \url{https://arxiv.org/abs/2504.03926}.

\bibitem[Jun and Zhang(2020)]{jun2020crop}
Kwang-Sung Jun and Chicheng Zhang.
\newblock Crush Optimism with Pessimism: Structured Bandits Beyond Asymptotic Optimality.
\newblock In \emph{Advances in Neural Information Processing Systems}, volume~33, pp.\ 6366--6376, 2020.
\url{https://arxiv.org/abs/2006.08754}.

\bibitem[Kannan et~al.(2018)Kannan, Morgenstern, Roth, Waggoner, and Wu]{kannan2018smoothed}
Sampath Kannan, Jamie~H. Morgenstern, Aaron Roth, Bo Waggoner, and Zhiwei Steven Wu.
\newblock A smoothed analysis of the greedy algorithm for the linear contextual bandit problem.
\newblock In \emph{Advances in Neural Information Processing Systems}, volume~31, 2018.

\bibitem[Kveton et~al.(2015)Kveton, Wen, Ashkan, and Szepesv\'ari]{kveton2015tight}
Branislav Kveton, Zheng Wen, Azin Ashkan, and Csaba Szepesv\'ari.
\newblock Tight regret bounds for stochastic combinatorial semi-bandits.
\newblock In \emph{Proceedings of the Eighteenth International Conference on Artificial Intelligence and Statistics}, volume 38, pp. 535--543. PMLR, 2015.
\newblock URL \url{https://proceedings.mlr.press/v38/kveton15.html}.

\bibitem[Oh et~al.(2019)Oh, Appavoo, and Gilbert]{oh2019periodic}
Shunhao Oh, Anuja Meetoo Appavoo, and Seth Gilbert.
\newblock Periodic bandits and wireless network selection.
\newblock In \emph{46th International Colloquium on Automata, Languages, and Programming}, pp. 149:1--149:15, 2019.
\newblock URL \url{https://arxiv.org/abs/1904.12355}.

\bibitem[Ortner et~al.(2012)Ortner, Ryabko, Auer, and Munos]{ortner2012restless}
Ronald Ortner, Daniil Ryabko, Peter Auer, and R\'emi Munos.
\newblock Regret bounds for restless Markov bandits.
\newblock In \emph{Algorithmic Learning Theory}, volume 7568, pp. 214--228. Springer, 2012.
\newblock doi: 10.1007/978-3-642-34106-9\_19.

\bibitem[Raghavan et~al.(2018)Raghavan, Slivkins, Wortman~Vaughan, and Wu]{raghavan2018diversity}
Manish Raghavan, Aleksandrs Slivkins, Jennifer Wortman Vaughan, and Zhiwei Steven Wu.
\newblock The externalities of exploration and how data diversity helps exploitation.
\newblock In \emph{Proceedings of the 31st Conference on Learning Theory}, volume~75 of \emph{Proceedings of Machine Learning Research}, pp.\ 1724--1738. PMLR, 2018.
\newblock URL \url{https://proceedings.mlr.press/v75/raghavan18a.html}.

\bibitem[Rife and Boorstyn(1974)]{rife1974single}
D.~C. Rife and R.~R. Boorstyn.
\newblock Single tone parameter estimation from discrete-time observations.
\newblock \emph{IEEE Transactions on Information Theory}, 20(5):591--598, 1974.
\newblock doi: 10.1109/TIT.1974.1055282.

\bibitem[Russo et~al.(2018)Russo, Van Roy, Kazerouni, Osband, and Wen]{russo2018tutorial}
Daniel Russo, Benjamin Van Roy, Abbas Kazerouni, Ian Osband, and Zheng Wen.
\newblock A tutorial on Thompson sampling.
\newblock \emph{Foundations and Trends in Machine Learning}, 11(1):1--96, 2018.
\newblock doi: 10.1561/2200000070.
\newblock URL \url{https://arxiv.org/abs/1707.02038}.

\bibitem[Slivkins et~al.(2025)Slivkins, Xu, and Zuo]{slivkins2025greedy}
Aleksandrs Slivkins, Yunzong Xu, and Shiliang Zuo.
\newblock Greedy algorithms for structured bandits: A sharp characterization of asymptotic success / failure.
\newblock In \emph{Advances in Neural Information Processing Systems}, volume~38, pp.\ 135156--135195, 2025.
\newblock doi: 10.52202/085713-4510.

\bibitem[Vovk(1998)]{vovk1998regression}
Volodya Vovk.
\newblock Competitive on-line linear regression.
\newblock In \emph{Advances in Neural Information Processing Systems}, volume 10, pp. 364--370, 1998.
\newblock URL \url{https://proceedings.neurips.cc/paper_files/paper/1997/file/30c8e1ca872524fbf7ea5c519ca397ee-Paper.pdf}.

\bibitem[Vovk(2001)]{vovk2001statistics}
Volodya Vovk.
\newblock Competitive on-line statistics.
\newblock \emph{International Statistical Review}, 69(2):213--248, 2001.
\newblock doi: 10.1111/j.1751-5823.2001.tb00457.x.

\bibitem[Wasserman et~al.(2020)Wasserman, Ramdas, and Balakrishnan]{wasserman2020universal}
Larry Wasserman, Aaditya Ramdas, and Sivaraman Balakrishnan.
\newblock Universal inference.
\newblock \emph{Proceedings of the National Academy of Sciences}, 117(29):16880--16890, 2020.
\newblock doi: 10.1073/pnas.1922664117.

\bibitem[Whittle(1988)]{whittle1988restless}
Peter Whittle.
\newblock Restless bandits: Activity allocation in a changing world.
\newblock \emph{Journal of Applied Probability}, 25(A):287--298, 1988.
\newblock doi: 10.2307/3214163.

\end{thebibliography}
\end{document}